\documentclass[letterpaper]{article} 
\usepackage{aaai2026}  
\usepackage{times}  
\usepackage{helvet}  
\usepackage{courier}  
\usepackage[hyphens]{url}  
\usepackage{graphicx} 
\usepackage{natbib}  
\usepackage{caption} 
\usepackage{algorithm}
\usepackage{algorithmic}
\usepackage{amsfonts} 
\usepackage{amsmath} 
\usepackage{subcaption}
\usepackage{booktabs}
\usepackage{comment}
\usepackage{graphicx} 
\usepackage{multirow} 
\usepackage{pifont}

\usepackage{threeparttable}
\usepackage{makecell}
\newcommand{\cmark}{\ding{51}} 
\newcommand{\xmark}{\ding{55}} 

\usepackage{newfloat}
\usepackage{listings}
\DeclareCaptionStyle{ruled}{labelfont=normalfont,labelsep=colon,strut=off} 
\floatstyle{ruled}
\newfloat{listing}{tb}{lst}{}
\floatname{listing}{Listing}

\usepackage{xcolor}

\usepackage[most]{tcolorbox}
\tcbset{
  breakable,          
  enhanced,
  colback=gray!10,
  colframe=black!75,
  boxrule=0.5pt,
  arc=2mm
}

\lstdefinelanguage{json}{
    basicstyle=\ttfamily\small,
    numbers=left,
    numberstyle=\tiny\color{gray},
    stepnumber=1,
    numbersep=8pt,
    showstringspaces=false,
    breaklines=true,
    frame=single,
    backgroundcolor=\color{gray!10},
    literate=
     *{0}{{{\color{blue}0}}}{1}
      {1}{{{\color{blue}1}}}{1}
      {2}{{{\color{blue}2}}}{1}
      {3}{{{\color{blue}3}}}{1}
      {4}{{{\color{blue}4}}}{1}
      {5}{{{\color{blue}5}}}{1}
      {6}{{{\color{blue}6}}}{1}
      {7}{{{\color{blue}7}}}{1}
      {8}{{{\color{blue}8}}}{1}
      {9}{{{\color{blue}9}}}{1}
      {:}{{{\color{black}:}}}{1}
      {,}{{{\color{black},}}}{1}
      {"}{{{\color{red}"}}}{1},
}
\title{TowerMind: A Tower Defence Game Learning Environment and Benchmark for LLM as Agents}
\author{
    Dawei Wang\textsuperscript{\rm 1},
    Chengming Zhou\textsuperscript{\rm 1},
    Di Zhao\textsuperscript{\rm 2},
    Xinyuan Liu\textsuperscript{\rm 1},
    Marci Chi Ma\textsuperscript{\rm 1},
    \\Gary Ushaw\textsuperscript{\rm 1},
    Richard Davison\textsuperscript{\rm 1}
}
\affiliations{
    \textsuperscript{\rm 1}Newcastle University, United Kingdom\\
    \textsuperscript{\rm 2}University of Auckland, New Zealand\\

    \{d.wang28, c.zhou10, x.liu89, c.ma20, gary.ushaw, richard-gordon.davison\}@newcastle.ac.uk, dzha866@aucklanduni.ac.nz
}

\usepackage{bibentry}

\begin{document}

\maketitle

\begin{abstract}
Recent breakthroughs in Large Language Models (LLMs) have positioned them as a promising paradigm for agents, with long-term planning and decision-making emerging as core general-purpose capabilities for adapting to diverse scenarios and tasks. Real-time strategy (RTS) games serve as an ideal testbed for evaluating these two capabilities, as their inherent gameplay requires both macro-level strategic planning and micro-level tactical adaptation and action execution. Existing RTS game-based environments either suffer from relatively high computational demands or lack support for textual observations, which has constrained the use of RTS games for LLM evaluation. Motivated by this, we present TowerMind, a novel environment grounded in the tower defense (TD) subgenre of RTS games. TowerMind preserves the key evaluation strengths of RTS games for assessing LLMs, while featuring low computational demands and a multimodal observation space, including pixel-based, textual, and structured game-state representations. In addition, TowerMind supports the evaluation of model hallucination and provides a high degree of customizability. We design five benchmark levels to evaluate several widely used LLMs under different multimodal input settings. The results reveal a clear performance gap between LLMs and human experts across both capability and hallucination dimensions. The experiments further highlight key limitations in LLM behavior, such as inadequate planning validation, a lack of multifinality in decision-making, and inefficient action use. We also evaluate two classic reinforcement learning algorithms: Ape-X DQN and PPO. By offering a lightweight and multimodal design, TowerMind complements the existing RTS game-based environment landscape and introduces a new benchmark for the AI agent field. The source code is publicly available on GitHub.\footnote{\url{https://github.com/tb6147877/TowerMind}}
\end{abstract}


\section{Introduction}

One of the fundamental challenges in artificial intelligence (AI) is equipping agents with the ability to solve tasks across a broader range of scenarios \cite{russell2016artificial}. Recent breakthroughs in large language models (LLMs) \cite{devlin2019bert, achiam2023gpt} have made them a promising approach to addressing this challenge. Benefiting from their extensive cross-domain knowledge and diverse abilities, including reasoning \cite{wei2022emergent, wang2022iteratively} and problem-solving \cite{lingo2024enhancing, renze2024self}, LLM-based agents have shown potential in various domains, such as healthcare \cite{li2024agent}, office automation \cite{zhang2024tablellm}, and design \cite{ccelen2024design}. Despite differences in context and specifics, these tasks consistently require two foundational capabilities from LLMs: \textbf{long-term planning} and \textbf{decision-making}, which are essential for accomplishing tasks: (1) LLMs leverage long-term planning to decompose a high-level task into a sequence of subgoals that guide progress toward the final objective; (2) LLMs perform decision-making to translate this sequence of subgoals into executable actions, conditioned on the evolving task state.

\begin{figure*}
    \centering
    \includegraphics[width=1.0\textwidth]{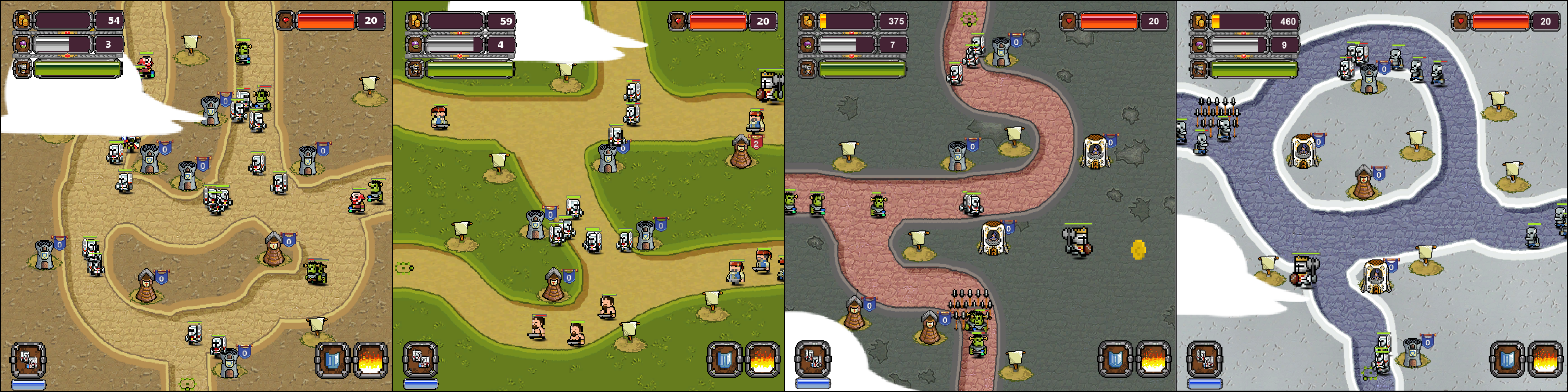}
    \caption{These are screenshots from four different TowerMind levels. The icons in the four corners of each image display key gameplay information, including the number of player’s current gold coins, player’s base health, and remaining enemy waves. The maps feature irregular, intersecting roads along which enemies advance toward the player's base in successive waves. Players must strategically build different types of towers at designated locations along these roads to repel the incoming enemies. The cloud-shaped white areas represent fog of war, introducing partial observability to the environment.}
    \label{fig:fig1}
\end{figure*}

Real-time strategy (RTS) games are an ideal platform for evaluating long-term planning and decision-making abilities, as they require players to engage simultaneously in both macromanagement and micromanagement \cite{barros2025deep}. Specifically, RTS games provide a battlefield setting where macromanagement tends toward long-term strategic planning, in which players formulate high-level strategies such as overall unit deployment and resource allocation; whereas micromanagement focuses on real-time decision-making, where players flexibly control units in response to dynamic changes on the battlefield to execute their combat plans. Currently, several RTS game-based benchmarks have recently been proposed for evaluating LLMs, including TextStarCraft II \cite{ma2025large}, LLM-PySC2 \cite{2024arXiv241105348L}, and VLMs Play StarCraft II \cite{ma2025vlms}, all of which are based on the StarCraft II Learning Environment (SC2LE) \cite{vinyals2017starcraft}, known for its relatively high computational demands. These challenging RTS game-based benchmarks are effective for assessing the long-term planning and decision-making capabilities of LLMs; nevertheless, the need for low-cost evaluation environments still persists in the field \cite{dubois2024length}. For example, in fast-paced continuous development pipelines \cite{koc2025tiny} and in the usage of reward models for instruction tuning \cite{yuan2024self}, lightweight benchmarks offer clear advantages. While several lightweight RTS game-based environments (e.g., ELF \cite{tian2017elf}, DeepRTS \cite{andersen2018deep}, Gym-$\mu$rts \cite{huang2021gym}) have been proposed to alleviate the computational demands of SC2LE-based platforms, they fundamentally lack support for textual observations and action interfaces, which makes them incompatible with LLMs.

To address the lack of lightweight RTS game-based environments with textual observation capabilities, we propose \textbf{TowerMind}, a newly developed game environment built upon the tower defense (TD) subgenre of RTS games \cite{liu2019automatic}, its screenshot is shown in Figure~\ref{fig:fig1}. TD games share the same core game mechanics as classic RTS games \cite{tian2017elf}, providing a battlefield scenario where players must build towers and deploy units to defend against waves of invading enemies, requiring them to demonstrate long-term planning and decision-making. Unlike the player-versus-player mechanics of classic RTS games, TD games focus solely on defending against predefined waves of enemies. This allows for a more isolated evaluation of LLMs’ ability to finish complex tasks using long-term planning and decision-making, without interference from opponent unpredictability. Furthermore, the fixed tower placement options and predefined enemy roads in TD games facilitate clearer analysis of the strategies employed by LLMs. In this work, TowerMind significantly reduces the computational demands compared to existing RTS game-based LLM benchmarks. Specifically, existing RTS game-based benchmarks for LLMs rely on the SC2LE environment, which requires approximately 30 GB of disk space, 2 GB of RAM, and a dedicated GPU. In contrast, TowerMind requires only 0.15 GB of disk space and RAM, runs efficiently on CPUs without the need for a dedicated GPU, and additionally offers advantages in ease of deployment and integration. This makes it well-suited for rapid research iteration, large-scale parallel training or fine-tuning, and similar scenarios in the LLM domain \cite{peng2023instruction}. Meanwhile, TowerMind supports pixel-based, textual, and structured game-state observations, enabling evaluation of multimodal LLMs. A comprehensive comparison of TowerMind and other lightweight RTS game-based environments in terms of supported features is provided in Table~1.

\begin{table*}[ht]
  
  \label{tab:compare_lightweight}
  \centering
\fontsize{9pt}{9pt}\selectfont
\setlength{\tabcolsep}{1.3mm}
  \begin{tabular}{lccccccc}
    \toprule
    \textbf{Environment} &  \makecell{\textbf{Pixel}\\ \textbf{Observation} } & \makecell{\textbf{Textual}\\ \textbf{Observation}}  & \makecell{\textbf{Stochastic}\\ \textbf{Environments}} & \makecell{\textbf{Partial}\\ \textbf{Observability}}& \makecell{\textbf{Level}\\ \textbf{Editor}}& \makecell{\textbf{Gym}\\ \textbf{Interface}} \\
    \midrule
    ELF \cite{tian2017elf}      & \cmark & \xmark      & \cmark & \cmark & \xmark & \xmark\\
    DeepRTS \cite{andersen2018deep}               & \cmark & \xmark      & \xmark & \cmark & \xmark& \xmark \\
    Gym-$\mu$rts \cite{huang2021gym}         & \xmark & \xmark     & \cmark & \cmark & \xmark & \cmark\\
    Mini HoK \cite{liu2024mini} & \xmark & \xmark     & \cmark & \xmark & \cmark& \xmark \\
    \midrule
    \textbf{TowerMind (Ours)}          & \cmark & \cmark     & \cmark & \cmark & \cmark& \cmark \\
    \bottomrule
  \end{tabular}
  \caption{Comparison between TowerMind and other lightweight RTS game-based environments.}
\end{table*}

In addition to addressing the limitations of existing RTS game-based environments, the design of TowerMind incorporates two new features: \textbf{(1) Hallucination Evaluation}: In evaluating LLMs, our metrics consider not only in-game score as a measure of performance, but also the executability of actions as an indicator of \emph{hallucination}. Hallucination refers to LLM outputs that conflict with factual or contextual information \cite{bang2023multitask}; in our setting, this specifically denotes actions that are invalid or inconsistent with the game state or rules. Such a metric design allows for simultaneous evaluation of LLM capabilities and reliability; \textbf{(2) Customizability}: As both a TD environment and engine, TowerMind includes a graphical level editor that enables researchers to conveniently create custom levels. These levels can range from trivially easy to extremely difficult or structurally unique, supporting diverse research needs and reducing the risk of data contamination.

The contributions of our work are four-fold: (1) We present TowerMind, a lightweight and multimodal TD environment for evaluating long-term planning and decision-making in LLMs, while also supporting hallucination analysis and offering strong customizability.  (2) The evaluation results show that while commercial LLMs outperform open-source ones, there remains a significant gap between LLMs and human experts. Furthermore, we observe several behavioural shortcomings during evaluation, including inadequate planning validation, a lack of multifinality in decision-making, and inefficient action use. (3) Based on the experimental results, we discuss three key aspects: how visual input enhances LLM capabilities, how correctness relates to effectiveness, and how LLMs handle misleading information. These findings provide insights for future research and highlight TowerMind’s potential as a versatile benchmark. (4) We evaluate two popular RL algorithms, Ape-X DQN \cite{horgan2018distributed} and PPO \cite{schulman2017proximal}, and the results demonstrate that TowerMind is a challenging environment that broadens the diversity of RL benchmarks.

\section{Related Work}

\textbf{Long-Term Planning and Decision-Making with LLMs.} Recent advances in LLMs have sparked growing interest in their capabilities beyond text generation, particularly in long-term planning and decision-making tasks \cite{brown2020language}. ReAct \cite{yao2023react} and Toolformer \cite{schick2023toolformer} demonstrate that LLMs can be guided to plan and act through reasoning traces and tool use, respectively. More recent approaches such as AutoGPT \cite{yang2023auto} and BabyAGI \cite{calegario2023exploring} attempt to leverage LLMs in open-ended task planning by forming feedback loops that allow the models to iteratively set subgoals and update plans. Whether originating from LLMs themselves or enabled through prompt engineering or agentic systems, long-term planning and decision-making capabilities need to be systematically evaluated to support further improvement. Different benchmarks adopt various perspectives when evaluating long-term planning and decision-making. PlanGenLLMs \cite{wei2025plangenllms}, AGENTBENCH \cite{liu2023agentbench}, and PLANET \cite{li2025planet} focus on simulated operating systems, web tasks, and other interactive environments. TowerMind, by leveraging the TD genre as a subclass of RTS games, preserves the strengths of RTS-style evaluation while providing a more computationally efficient alternative.

\vspace{0.3em} 

\noindent \textbf{RTS Game-Based Environments and Benchmarks}. RTS games have long been one of the game genres most deeply intertwined with AI research \cite{buro2003real}. Currently, available RTS game-based environments include SC2LE \cite{vinyals2017starcraft}, StarCraft Multi-Agent Challenge (SMAC) \cite{samvelyan2019starcraft}, Honor of Kings Arena and Honor of Kings 3v3 Arena (HoK3v3) \cite{wei2022honor}, along with several lightweight alternatives designed for lower computational demands, such as ELF \cite{tian2017elf}, DeepRTS \cite{andersen2018deep}, Gym-$\mu$rts \cite{huang2021gym} and Mini Honor of Kings (Mini HoK) \cite{liu2024mini}. To facilitate LLM-related research, several RTS game benchmarks featuring text-based observations have been introduced, such as TextStarCraft II \cite{ma2025large}, LLM-PySC2 \cite{2024arXiv241105348L}, and VLMs Play StarCraft II \cite{ma2025vlms}, all of which face relatively high computational demands because they are based on SC2LE. TowerMind serves as a computationally efficient alternative to existing RTS game-based benchmarks for LLM evaluation. Moreover, it extends beyond current lightweight RTS game-based environments by supporting multimodal observations and functionalities that reflect the evolving needs of AI agent research.

\vspace{0.3em} 

\noindent \textbf{Tower Defense Games and AI Research.} TD games represent a rule-convergent subgenre of RTS games, characterized by a combination of simple mechanics and substantial strategic depth \cite{5949738}. With the advancement of AI, TD games have increasingly attracted attention in AI research. Early work focused on applying AI algorithms to address TD-specific challenges \cite{rummell2011adaptive, tan2013automated, wong2015game}, while more recent studies have adopted TD games as testbeds for reinforcement learning (RL) research \cite{dias2020reinforcement, bergdahl2024reinforcement} and human-AI collaboration \cite{haduong2024cps}. However, the only TD game available to the research community is a small module in the ELF \cite{tian2017elf} environment, which is highly limited in tower and enemy variety and lacks units control. To date, the research community still lacks a dedicated and sufficiently challenging TD game environment. TowerMind fills this gap and enriches the diversity of benchmarks in the AI agent research domain.

\section{The TowerMind Environment}
The TowerMind environment is built on top of the Unity game engine and extended into an AI environment using the Unity ML-Agents Toolkit \cite{juliani2018unity}.

\subsection{Game Mechanics}
\label{sec:game_mechanics}

TowerMind consists of a series of independent levels, each with a distinct map and enemy configuration, where enemies spawn in sequential waves and advance toward the player's base. Players must strategically build towers and control units to prevent enemies from reaching the base and depleting the player's base health. For more details about the game mechanics, please refer to the Appendix~\ref{appendix:game_mechanics_details}.

\vspace{0.3em} 

\begin{figure*}
    \centering
    \includegraphics[width=0.8\textwidth]{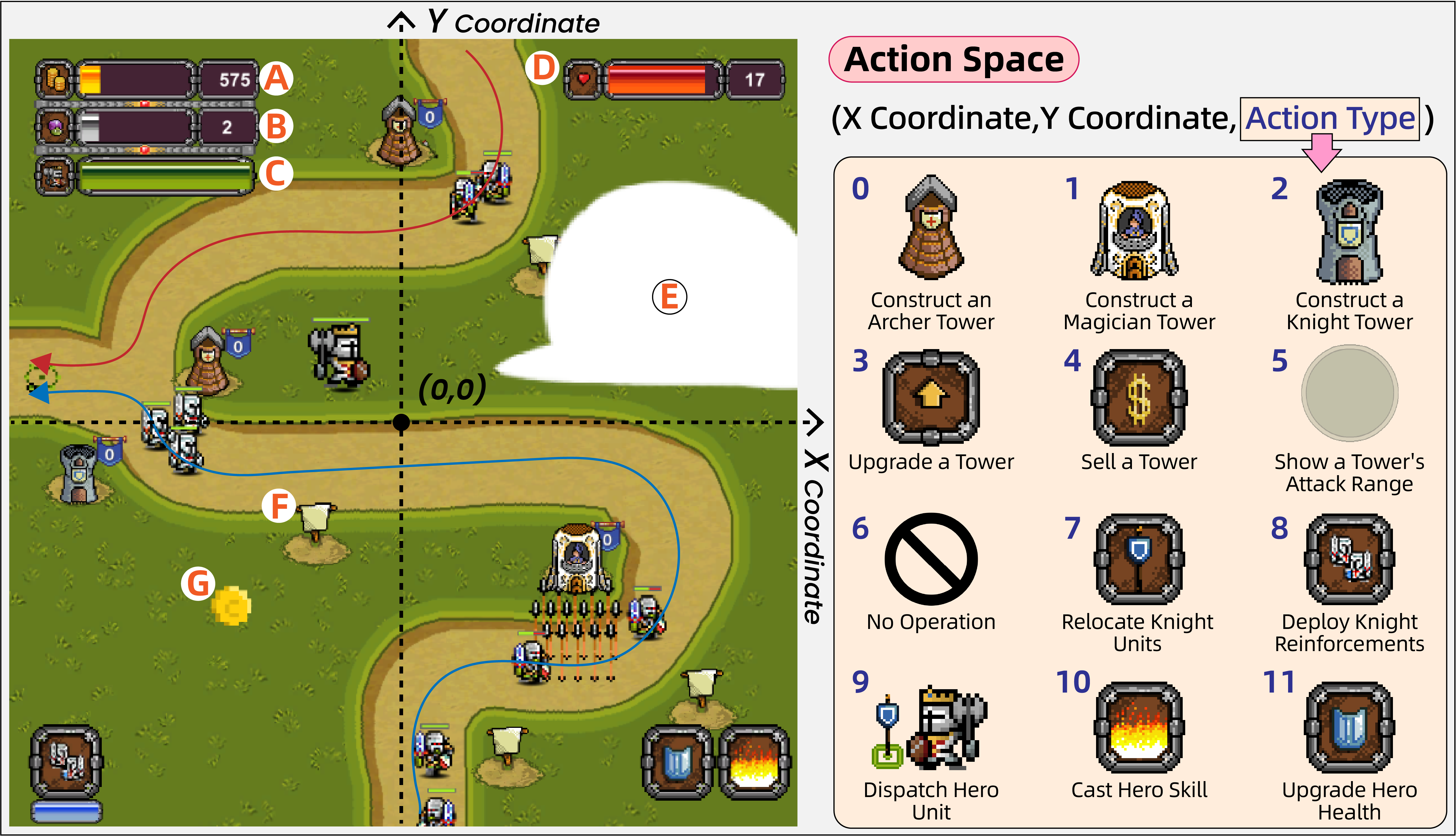}
    \caption{Left: A screenshot of the TowerMind environment with key game elements annotated. The coordinate axes illustrate the alignment between the game map and the 2D coordinate system. The red and blue arrowed curves represent the two roads used by enemies to attack. Labels A–G indicate: (A) player's current gold coins; (B) remaining enemy waves; (C) hero unit's current health; (D) player's base health; (E) fog of war; (F) an unbuilt tower point; and (G) gold coins dropped on the map awaiting collection; (H) knights units; (I) an enemy. Right: Illustrations and brief descriptions of the 12 action types.}
    \label{fig:fig2}
\end{figure*}

\noindent \textbf{Maps.} In TowerMind, the map is defined as a square area with a side length of 6, centered at $(0.0, 0.0)$,  where both the horizontal and vertical coordinates lie within the interval $[-3.0, 3.0]$. It consists of two fundamental elements: roads and tower points. Roads (the red and blue directional curves in Figure~\ref{fig:fig2}) are fixed paths that guide enemy movement through the map, represented as sequences of 2D coordinate waypoints along which enemies move in straight lines, often starting from different locations and converging at the player's base. Tower points (the label "F" in Figure~\ref{fig:fig2}) are predefined locations on the map where players can construct defensive towers, typically positioned along both sides of the roads. Some tower points, however, are placed far from the roads and cannot engage any enemies, thus serving as misleading tower points. Together, the varying shapes of roads and the diverse locations of tower points form rich and dynamic map features, serving as a critical factor influencing players’ long-term planning and decision-making.

\vspace{0.3em} 

\noindent \textbf{Towers, Knights and Hero.} Players can control three types of game entities to defend against enemy attacks: towers, knight units, and a hero unit, with all player actions related to these three entities. There are three types of buildable towers: Archer Tower, Magician Tower, and Knight Tower, each with distinct construction costs, attack styles, and optimal use cases. The Archer Tower deals strong single-target damage, the Magician Tower performs area-of-effect (AoE) attacks, and the Knight Tower summons knight units that can be manually controlled by the player. Knight units  (the label ”H” in Figure 2)  are melee fighters designed for individual combat with enemy units. They can either be summoned from knight towers or directly deployed anywhere on the map via the knight reinforcements action. And in each level, players can control a hero unit with greater health, attack damage, and other attributes than knight units, and can manipulate its movement and skill usage with finer granularity. Overall, towers are relatively static and reflect high-level strategic planning, whereas knight units and the hero unit are more flexible and emphasize low-level tactical decision-making and action execution.

\vspace{0.3em}

\noindent \textbf{Enemies.} Enemies (the label ”I” in Figure 2)  appear in waves, with each wave consisting of a predefined number of enemies and a fixed time interval between waves. There are 15 distinct enemy types, each exhibiting different attributes in terms of health, movement speed, and attack damage. Some also possess special abilities; for example, the Orc Sorcerer can disable nearby towers. By varying the types and quantities of enemies within each wave, the game generates diverse enemy compositions, making it impossible for players to rely on a single fixed strategy to clear all levels.

\vspace{0.3em}

\noindent \textbf{Resources.} Gold coins (the label "A" in Figure~\ref{fig:fig2}) serve as the sole in-game resource and are required for constructing towers, upgrading existing towers, and enhancing the hero’s maximum health. Gold coins periodically appear at random locations on the map (the label "G" in Figure~\ref{fig:fig2}) and must be actively collected by dispatching knight units or the hero to the corresponding positions. Moreover, when the hero’s AoE skill inadvertently eliminates friendly knight units, a Friendly Fire Compensation mechanism is triggered, awarding the player an amount of gold coins as compensation.

\vspace{0.3em}

\noindent \textbf{Fog of War.} In each level, a white, cloud-shaped fog of war region is present and moves randomly across the map (the label "E" in Figure~\ref{fig:fig2}), introducing partial observability to the environment. All towers, knight units, the hero unit, and enemies located within the fog of war are excluded from the observation space, and friendly units in these areas become inactive and do not attack enemies. Fog of war increases the difficulty and uncertainty of the environment. 

\subsection{Environment Interface}
The TowerMind environment conforms to the OpenAI Gym standard \cite{brockman2016openai} for easy integration with existing frameworks. For more details, see Appendix~\ref{appendix:env_interface_details}.

\noindent \textbf{Observations.} TowerMind provides three distinct types of observations: pixel-based, textual, and structured game-state observations. The pixel-based observation is a $512 \times 512 \times 3$ colour image representing the raw game screen. The textual and structured game-state observations represent two different formats derived from the same underlying game-state information, which includes both level-specific and real-time status data. For instance, \texttt{Level\_Initial\_Gold\_Coins} indicates the initial number of gold coins at the beginning of a level, while \texttt{Level\_Enemies\_Realtime\_Status} provides real-time data about each enemy, such as their health points and coordinates. The textual observation explicitly encodes this information in JSON format, clearly associating each piece of data with its corresponding semantic field name, facilitating comprehension by LLMs. In contrast, the structured game-state observation presents the same information flattened into a one-dimensional array.

\noindent \textbf{Actions.} The action space in this environment is designed as a \emph{hybrid action space}, where each action $a$ is represented by a three-dimensional vector: $ a = (x, y, c) $. Here, the first two dimensions $(x,y)$ are continuous variables specifying the spatial location of the action, expressed as Cartesian coordinates within a two-dimensional plane centered at the map's midpoint, with both horizontal and vertical coordinates constrained by: $x, y \in [-3.0, 3.0]$, consistent with the spatial boundaries of the game map. The third dimension $c$ denotes the discrete action type (e.g., upgrade a tower, sell a tower, dispatch knight reinforcements), represented by an integer index: $c \in \{0, 1, 2, \dots, 11\}$. This action space design integrates continuous spatial coordinates with discrete action selections, enabling diverse interactions within the environment. Figure~\ref{fig:fig2} provides a more intuitive illustration of the action space. Additionally, only actions that comply with the game rules and current state are considered executable, and we refer to these as \textit{valid actions}. In contrast, actions that violate these constraints, such as issuing a tower-building command at coordinates where no tower point exists or when insufficient gold is available, are classified as invalid and are not executed.

\noindent \textbf{Reward.} TowerMind provides a sparse reward signal, assigning a reward of $-1.0$ for each enemy that reaches the player’s base, which aligns with the game mechanic where the player's base health (the label "D" in Figure~\ref{fig:fig2}) is reduced by one in the same situation.

\noindent \textbf{Episode Dynamics.} Each level in TowerMind is treated as a single finite episode, terminating either when the player's base health reaches zero or when all enemy waves have been eliminated. In all our experiments, actions are executed by default every 16 game steps, corresponding to 187 actions per minute, similar to the action frequency in SC2LE.

\subsection{Levels and Difficulty}
\label{sec:level_diff_metr}
 TowerMind includes five built-in benchmark levels with increasing difficulty from Level 1 to Level 5. To characterize each level, we propose a quantitative metric system grounded in the game mechanics described in Section~\ref{sec:game_mechanics}, enabling the modeling and comparison of level difficulty. We define the difficulty of a level $l$ in TowerMind as a scalar value $D(l)$, composed of four components: $D(l) = d_{r}(l) + d_{t}(l) + d_{e}(l) + d_{re}(l)$.

\noindent \textbf{Road}: $d_{r}(l) = \frac{R_l}{R_{\max}}$, where $R_l$ is the number of roads in level $l$, and $R_{max}$ is the design-time maximum number of roads in the TowerMind environment.

\noindent \textbf{Tower}: $d_{t}(l) = \frac{T_l}{T_{\max}}$, where $T_l$ is the number of tower points in level $l$, and $T_{max}$ is the design-time maximum number of tower points in the TowerMind environment.

\noindent \textbf{Enemy}: $d_{e}(l) = \frac{E_l}{E_{\text{total}}} + \frac{\bar{N}l}{N{\max}}$, where $E_l$ is the number of enemy types in level $l$, and $E_{total}$ is the total number of enemy types, $\bar{N}_l$ is the average number of enemies per wave in level $l$, and $N_{max}$ is the design-time maximum number of enemies per wave in the TowerMind environment.

\noindent \textbf{Resource}: $d_{re}(l) = \frac{1}{3} \left( \frac{I_{\min}}{I_l} + \frac{G_{\min}}{G_l} + \left(1 - r_{\text{sellback}}(l)\right)\right)$, where $I_{min}$ is the design-time minimum initial gold coins in this environment, $I_l$ is the initial gold coins in level $l$, $G_{min}$ is the design-time minimum gold coins drop in the TowerMind environment, $G_l$ is the gold coins drop amount in level $l$, and $r_{\text{sellback}}(l)$ is the tower sell-back ratio in level $l$.

\subsection{Environment Customizability}
The customizability of TowerMind can be categorized into three aspects: (1) \textbf{Level Customization}, researchers can easily create new levels using the provided graphical user interface-based level editor; (2) \textbf{Parameter Customization}, we have made almost all game-related parameters modifiable, allowing researchers to freely adjust attributes of towers, enemies, heroes, and other game elements; (3) \textbf{Feature Customization}, researchers can selectively enable or disable various modes and features within TowerMind, such as enabling debugging mode or activating human trajectory recording functionality. See Appendix~\ref{appendix:env_cus} for more details.

\section{Evaluation for LLMs}

\label{sec:llm_bench}

\subsection{Evaluation Setting}

We define two evaluation metrics: \textbf{score} and \textbf{valid action rate}. The score metric is identical to the raw reward signal provided by the TowerMind environment interface. As all benchmark levels assign the player base a fixed health of 20, the score metric is a real number ranging from –20 to 0. The valid action rate metric is calculated as the ratio of valid actions to total actions within a given level: $\frac{\# \text{Valid Actions}}{\# \text{Total Actions}}$, ranges from 0 to 1.

Our evaluation covers a range of popular commercial and open-source models, including GPT-4.1 \cite{achiam2023gpt}, Gemini-2.5-Pro \cite{comanici2025gemini}, Claude 3.7 Sonnet \cite{Anthropic}, Llama 3.2 (90B/11B) \cite{MetaAI}, and Qwen2.5-VL (72B/7B) \cite{bai2025qwen2}. We employ a zero-shot prompting strategy, using identical prompts across all models to ensure fairness and consistency. The language-only modality receives only the prompt, whereas the vision-language modality includes both the prompt and a $512 \times 512 \times 3$ pixel-based observation. To support a comprehensive benchmark, we evaluated the performance of five human experts across five benchmark levels to establish a human experts baseline. See Appendix~\ref{appendix:additional_experimental_results} for more details.

\subsection{Results}
In our experiments, each model was evaluated using five random seeds across each benchmark level under both language-only and vision-language modalities. Tables 2 and 3 respectively present the score and valid action rate performances of different LLMs on each benchmark level. All values in these tables are normalized relative to the human experts baseline. \textbf{Bold} values indicate the best-performing results on each benchmark level under the language-only modality, \underline{underlined} values denote the best-performing results on each benchmark level under the vision-language modality, and values highlighted in \textit{italic} indicate performance worse than the random baseline. Further tables including standard errors and additional experimental information are provided in Appendix~\ref{appendix:additional_experimental_results}.

\begin{table}[ht]
  
  \label{tab:reward-performance}
  \centering
  \fontsize{9pt}{9pt}\selectfont
  \setlength{\tabcolsep}{1.5mm}
  \begin{tabular}{l*{9}{c}c}
    \toprule
    \textbf{Model} &\textbf{Lv 1} & \textbf{Lv 2} & \textbf{Lv 3} & \textbf{Lv 4} & \textbf{Lv 5} & \textbf{Avg.} \\
    \midrule
    \multicolumn{7}{l}{\textbf{Language-Only}} \\
    GPT-4.1                      & $0.59 $ & $0.49 $ & $0.32 $ & $0.19 $ & $0.07 $ & $0.33 $ \\
    Gemini-2.5-Pro              & $0.52$  & $0.42 $ & $0.31 $ & $0.11 $ & $0.01 $ & $0.27 $ \\
    Claude 3.7 Sonnet          & \textbf{0.62}  & \textbf{0.51}  & \textbf{0.40}  & \textbf{0.24} &\textbf{0.15} & \textbf{0.38} \\
    Llama 3.2 90B               & $0.42  $& $0.32  $& $0.19  $& $0.12  $& 0.00  & $0.21  $ \\
    Llama 3.2 11B               & $0.17  $& $0.09  $& $0.00  $& $0.00  $& $0.00  $& $0.05  $ \\
    Qwen 2.5-VL 72B             & $0.47  $& $0.36  $& $0.21  $& $0.00  $& $0.00  $& $0.21 $ \\
    Qwen 2.5-VL 7B              & $0.00  $& $0.00  $& $0.00  $& $0.00  $& $0.00  $& $0.00  $ \\
    \midrule
    \multicolumn{7}{l}{\textbf{Vision-Language}} \\
    GPT-4.1               & $0.63 $& $0.56 $& $0.44 $& \underline{0.32}& $0.15 $& \underline{0.42} \\
    Gemini-2.5-Pro        & $0.57 $& $0.44 $& $0.33 $& $0.16 $& $0.01 $& $0.30 $ \\
    Claude 3.7 Sonnet    & \underline{0.67}& \underline{0.58}& \underline{0.45}& $0.20 $& \underline{0.16}& $0.41 $ \\
    Llama 3.2 90B         & $0.30 $& $0.05 $& $0.00 $& $0.00 $& $0.00 $& $0.07  $ \\
    Llama 3.2 11B         & $0.04 $& $0.00 $& $0.00 $& $0.00 $& $0.00 $& $0.01 $ \\
    Qwen 2.5-VL 72B       & $0.54 $& $0.39 $& $0.20 $& $0.12 $& $0.05 $& $0.26 $ \\
    Qwen 2.5-VL 7B        & $0.05 $& $0.00 $& $0.00 $& $0.00 $& $0.00 $& $0.01 $ \\
    \midrule
    Random                     &$0.00 $&$0.00 $&$0.00 $&$0.00 $&$0.00 $& $0.00 $                                               \\
    \bottomrule
  \end{tabular}
  \caption{The score performance of different LLMs on the benchmark levels.}
\end{table}

\begin{table}[ht]
  
  \label{tab:var-performance}
  \centering
  \fontsize{9pt}{9pt}\selectfont
   \setlength{\tabcolsep}{1.5mm}
  \begin{tabular}{l*{9}{c}c}
    \toprule
    \textbf{Model} & \textbf{Lv 1} & \textbf{Lv 2} & \textbf{Lv 3} & \textbf{Lv 4} & \textbf{Lv 5} & \textbf{Avg.} \\
    \midrule
    \multicolumn{7}{l}{\textbf{Language-Only}} \\
     GPT-4.1                     & \textbf{0.92} & $0.89 $ & $0.88 $ & $0.84 $ & $0.75 $ & $0.86 $ \\
    Gemini-2.5-Pro              & $0.91 $ & \textbf{0.90} & \textbf{0.89} & $0.83 $ & \textbf{0.82} & \textbf{0.87} \\
    Claude 3.7 Sonnet           & $0.90 $ & $0.87 $ & $0.85 $ & \textbf{0.85} & $0.79 $ & $0.85 $ \\
    Llama 3.2 90B              & $0.48 $& $0.39 $& $0.30 $& \textit{0.21}& $\textit{0.20}  $& $0.32  $ \\
    Llama 3.2 11B              & $0.28 $& \textit{0.24}& \textit{0.23}& \textit{0.23}& $\textit{0.22} $& $0.24 $ \\
    Qwen 2.5-VL 72B             & $0.87 $& $0.78 $& $0.76 $& $0.58 $& $0.51 $& $0.70 $ \\
    Qwen 2.5-VL 7B               & \textit{0.11} & \textit{0.05}& \textit{0.03}& \textit{0.01}& \textit{0.01}& \textit{0.04} \\

    \midrule
    \multicolumn{7}{l}{\textbf{Vision-Language}} \\
     GPT-4.1               & \underline{0.86}& $0.81 $& $0.75 $& $0.68 $& $0.66 $& $0.75 $ \\
    Gemini-2.5-Pro       & $0.85 $& $0.81 $& $0.80 $& $0.73 $& $0.67 $& $0.77 $ \\
    Claude 3.7 Sonnet    & $0.85 $& \underline{0.85}& \underline{0.83}& \underline{0.80}& \underline{0.79}& \underline{0.82} \\
    Llama 3.2 90B        & $0.44 $& $0.38 $& $0.33 $& $0.31 $& $0.30 $& $0.35 $ \\
    Llama 3.2 11B         & 0.31& $\textit{0.19} $& \textit{0.18}&\textit{0.13}& \textit{0.11}& \textit{0.18} \\
    Qwen 2.5-VL 72B      & $0.79 $& $0.72 $& $0.66 $& $0.54 $& $0.43 $& $0.63 $ \\
    Qwen 2.5-VL 7B        & \textit{0.21}& \textit{0.15}& \textit{0.05}& \textit{0.04}& \textit{0.02}& \textit{0.09} \\

    \midrule
    Random &$0.25 $&$0.25 $&$0.24 $&$0.24 $&$0.22 $&$0.24 $\\

    \bottomrule
  \end{tabular}
  \caption{The valid action rate performance of different LLMs on the benchmark levels.}
\end{table}

\subsection{Quantitative Analysis}
Based on the data in Tables 2 and 3, including both individual benchmark level results and overall averages, we identify the following key findings:

\noindent \textbf{Limited Performance of LLMs.} In terms of score, Claude 3.7 Sonnet achieved the best performance in the language-only setting, and GPT-4.1 in the vision-language setting. However, they still lag behind human experts by 62\% and 58\%, with all other models showing even larger gaps. Notably, on the most difficult level, Level 5, all models underperform human experts by at least 84\%.

\noindent \textbf{Vision Input Improves Performance.} All evaluated models except Llama 3.2 (90B/11B) show improved score performance under the vision-language modality compared to the language-only modality. This suggests that multimodal cues enhance these models' environmental understanding, whereas Llama 3.2 (90B/11B) seems to struggle with such complex and dynamic visual inputs.

\noindent \textbf{Hallucination Issues in Open-Source LLMs.} From the perspective of valid action rate, the three commercial LLMs performed relatively well, each exhibiting a gap of less than 20\% compared to human experts. Among the open-source models, only Qwen 2.5-VL 72B showed acceptable results, while the other three models underperformed significantly. In particular, the smaller models Qwen 2.5-VL 7B and Llama 3.2 11B exhibited performance on several benchmark levels that was even lower than the random baseline. The high level of hallucination constrains their long-term planning and decision-making capabilities.

\noindent \textbf{Effect of Level Difficulty on Hallucination.} As level difficulty increases, the degree of hallucination also rises across models. This suggests that harder levels tend to include more game elements, resulting in longer prompts that challenge the models' generation stability and consistency.

\subsection{Qualitative Analysis}
We analyzed model trajectories and identified common challenges and limitations:

\noindent \textbf{Insufficient Validation of Long-Term Planning.} In Levels 1 and 2, we placed one or more misleading tower points located far from the enemy attack roads. Building towers on these tower points does not threaten any enemies, thus serving only to waste resources. However, the LLMs consistently chose to build towers on these misleading tower points. Despite having access to all necessary information in the prompt to compute that these towers would not engage any enemies, the models failed to perform such basic spatial or numerical reasoning during tower placement planning.

\noindent \textbf{Decision-Making Without Multifinality Thinking.} Multifinality refers to the ability to achieve multiple goals with a single action \cite{kruglanski2015architecture}, and it is a key decision-making skill for optimizing task efficiency. This type of behavior is frequently observed in human expert gameplay. For example, human experts may direct the hero unit to collect gold coins while simultaneously attacking nearby enemies. However, we have never observed such behavior in the gameplay trajectories of any LLMs.

\noindent \textbf{Limited Use and Understanding of Actions.} We frequently observe that LLMs fail to fully utilize or understand the available action space. Typical behaviors include neglecting to upgrade towers despite sufficient gold, sending knight reinforcements to empty areas, or using the hero’s AoE skill in the absence of enemies.  This suggests that LLMs tend to interpret the available actions only at a surface level, lacking a deeper understanding of their strategic use and appropriate contexts.

\subsection{Insights and Future Directions}
Based on the above quantitative and qualitative analyses, we attempt to further discuss the findings in this section, with the hope of providing insights that may inspire future research in the LLM domain.

\noindent \textbf{Effect of Visual Input on Model Performance.} The majority of LLMs evaluated in our experiments exhibited improved performance under the vision-language modality compared to the language-only setting. This indicates that LLMs can leverage visual inputs to access information beyond what is conveyed through text, thereby enhancing their reasoning, planning, and decision-making capabilities. Consequently, future research may benefit from further exploring the role of visual information, such as through vision-informed prompt engineering or preprocessing techniques for visual feature extraction.

\noindent \textbf{From Correctness to Effectiveness.} In our experiments, we found that the gap between LLMs and human experts in terms of valid action rate is smaller than the gap in score. This suggests that LLMs are capable of understanding the game rules and current game state, and can generate actions that are consistent with both. However, these actions are often limited in their effectiveness toward achieving the intended goals. This is analogous to a common issue observed in LLM-based question answering, where models often produce 'technically correct but ultimately unhelpful' responses. Accordingly, evaluating LLMs in the future should go beyond static knowledge benchmarks like SuperGLUE \cite{wang2019superglue} and MMLU \cite{hendrycks2020measuring}, which test correctness, and increasingly incorporate interactive benchmarks such as AGENTBENCH \cite{liu2023agentbench} and TowerMind, which assess the effectiveness of model-generated responses in dynamic, decision-making settings.

\noindent \textbf{Identifying Misleading Content.} In our experiments, even the most advanced LLMs were observed to waste resources by building towers at misleading tower points that would never engage any enemies. The ability to identify misleading information is critical not only for improving the performance of LLMs, but also for ensuring their safety. LLMs need mechanisms to prevent being misled into producing toxic content \cite{bianchi2024large}. This highlights the necessity of incorporating validation mechanisms into LLM systems. Importantly, these mechanisms should be outcome-driven, meaning that validation should extend beyond surface-level textual content to include assessment of the effects or implications of the generated content.

\section{RL Benchmark}

\label{sec:rl_bench}
To validate TowerMind’s feasibility and challenge in RL, we established a preliminary RL benchmark by adopting two widely-used algorithms as baselines: Ape-X DQN~\cite{horgan2018distributed} and PPO~\cite{schulman2017proximal}. We trained two algorithms on the five benchmark levels using pixel-based and structured game-state observations. Each algorithm was run three times with different random seeds, using 100 million environment steps per run, as shown in Figure~\ref{fig:rl_dqn_vs_ppo}. We further evaluated the trained models on the benchmark levels using five different random seeds, following the same score metric used in the LLM-based evaluation, as shown in Figure~\ref{fig:rl_in_levels}. See Appendix~\ref{appendix:rl_experiment} for detailed information about the RL experiments.

\begin{figure}[t]
\centering
\includegraphics[width=0.9\columnwidth]{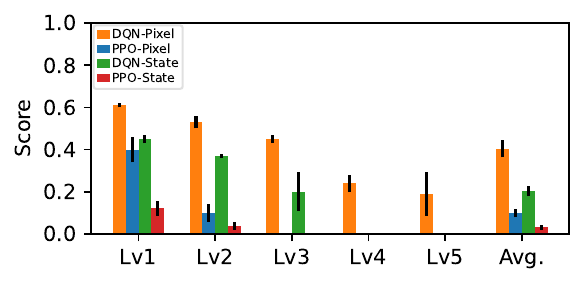} 
\caption{The evaluation results on the benchmark levels, with scores normalized relative to the human expert. Error bars represent the standard error.}
\label{fig:rl_in_levels}
\end{figure}

\begin{figure}[t]
\centering
\includegraphics[width=1.0\columnwidth]{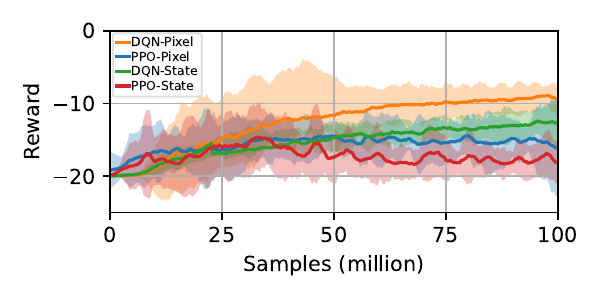} 
\caption{Training curves, the horizontal axis shows the number of training samples, measured in millions. Error bars represent 95\% confidence intervals.}
\label{fig:rl_dqn_vs_ppo}
\end{figure}

The evaluation results indicate that, after 100 million environment steps, both RL algorithms were able to solve simpler levels to some extent, but their performance remained substantially inferior to that of human experts, which suggests that TowerMind is a challenging environment for RL.

\section{Conclusion}
\label{sec:conclusion}
In this work, we propose TowerMind, a lightweight TD game environment with multimodal observation capabilities designed for evaluating LLMs. Through our evaluation, TowerMind reveals a substantial performance gap between current LLMs and human experts in terms of long-term planning and decision-making. It also clearly demonstrates the roles of visual input, hallucination, and misleading information in contributing to this gap, highlighting TowerMind's practical value for LLM research. Additionally, TowerMind can also be used for RL research. We believe that TowerMind can serve as a practical platform to facilitate the development of more capable AI agents.

\bibliography{main}


\clearpage

\appendix

\section{Game Mechanics Additional Details}
\label{appendix:game_mechanics_details}
A tower point is not a single coordinate, but a square region centered at that coordinate with a side length of 0.5. Any tower-related action such as building, upgrading, or selling a tower will be executed on a tower point if the action's coordinates fall within that tower point's square region.

In TowerMind, the Archer Tower can attack both flying and ground enemies, while the Magician Tower can only target ground enemies. The knight units summoned by the Knight Tower can also only attack ground enemies, the hero can attack both flying and ground enemies. Knight reinforcements can be deployed anywhere on the map, but are limited by a cooldown, which is 10 seconds by default. The hero is a versatile unit capable of auto-healing, dealing single-target damage and unleashing AoE skills by sacrificing its own health, but realizing its full potential requires precise and fine-grained control.

\begin{figure}
    \centering
    \includegraphics[width=1.0\linewidth]{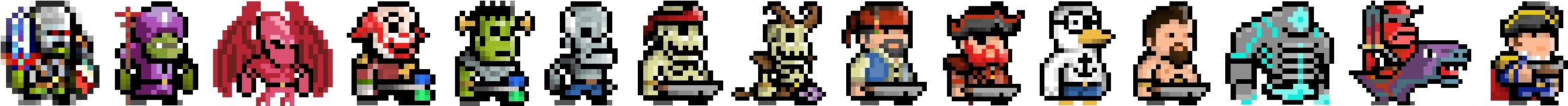}
    \caption{Screenshots of the 15 enemy types, with IDs from left to right ranging from 0 to 14.}
    \label{fig:enemies}
\end{figure}

Enemies are categorized as either flying or ground. Among the 15 enemy types, only the Demon Bat is a flying enemy; all others are ground enemies. The full list of 15 enemy types and
their properties is provided in below JSON configuration files and screenshots are provided in Figure~\ref{fig:enemies}.

In TowerMind, aside from gold coins, several other game elements can be considered resources in a broader sense: (1) the hero unit’s health can be viewed as a damage-convertible resource, since casting the hero’s AoE skill consumes health, which passively regenerates over time; (2) deploying knight reinforcements incurs no cost. By managing their cooldown effectively, players can maintain up to two additional knight units on the map to help defend against enemies. Thus, knight reinforcements can also be regarded as resources that can be converted into damage; (3) due to the Friendly Fire Compensation mechanism, players receive gold coins when the hero unit kills friendly knight units. Therefore, knight units can also be seen as resources that can be converted into gold coins.

Game elements covered by the Fog of War are removed from all forms of observation, including pixel-based, textual, and structured game-state observations. However, the Fog of War only inactivates friendly units, while enemies covered by it continue to move along the road. It is worth noting that players can temporarily illuminate the Fog of War by casting the hero’s flame-based AoE skill within its range, nullifying its effect for a short duration. 

All randomness in the game, such as damage calculation, gold coin collection, and enemy path selection, is entirely controlled by a user-defined integer seed. Given the same starting seed and the same sequence of actions, the environment produces identical outputs, ensuring the reproducibility for research purpose.

The following are the configuration tables for towers, knight units, the hero unit, knight reinforcements, and enemies in the environment, containing detailed data for each.

\begin{lstlisting}[caption=Configuration Table of Towers,language=json]
{"Towers":
[
  {"Type":1,"Price":100,"AttackSpeed":4.0, "AttackDamage":0,"AttackExtraDamage":0, "AttackRange": 2.0, "CanAttackAir":false, "CanAttackGround":false, "Name":"Knight Tower", "UpgradePrice":100, "UpgradeGrowth":1.2, "Description": "Knight Tower cannot directly attack the enemy, but it can summon up to three knights to the battlefield. After these knights die, the Knight Tower will continue to summon knights. You can deploy these three knights anywhere within the attack range of the Knight Tower."},
  {"Type":2,"Price":110,"AttackSpeed":2.0, "AttackDamage":100,"AttackExtraDamage":20, "AttackRange": 2.5, "CanAttackAir":false, "CanAttackGround":true, "Name":"Magician Tower","UpgradePrice":110, "UpgradeGrowth":1.3,"Description": "Magician Tower can only attack ground enemies, not air enemies. It can create a blade trap area (a square with a side length of 0.8) on the ground, any enemy passing through this area will be damaged. "},
  {"Type":3,"Price":120,"AttackSpeed":0.8, "AttackDamage":100,"AttackExtraDamage":50, "AttackRange": 3.0, "CanAttackAir":true, "CanAttackGround":true, "Name":"Archer Tower","UpgradePrice":120, "UpgradeGrowth":1.4,"Description": "Archer Tower can attack enemies on the ground and in the air. It can only hit one enemy at a time. "}
]}
\end{lstlisting}

\begin{lstlisting}[caption=Configuration Table of Knight Units,language=json]
{"FilePath":"Prefabs/Knight", "Health":600, "MovementSpeed":0.6,"AttackSpeed":0.7, "AttackDamage":150,"AttackExtraDamage": 50,"AttackRange": 1.0,"CanAttackAir":false, "CanAttackGround":true, "FFCompensationValue": 50, "FFCompensationProbability": 1.0, "Description": "Whether they are knights summoned by Knight Tower(s) or knight squads sent directly to the battlefield, all knights are within the scope of the 'Friendly Fire Compensation (FFCompensation)' mechanism."}
\end{lstlisting}

\begin{lstlisting}[caption=Configuration Table of The Hero Unit,language=json]
{"FilePath":"Prefabs/Hero", "Health":1600, "MovementSpeed":0.9,"AttackSpeed":0.7, "AttackDamage":200,"AttackExtraDamage": 150, "AttackRange": 1.0,"SkillAttackDamage": 100,"SkillAttackExtraDamage": 100,"SkillCostHealth": 100, "SkillLastTime": 5.0, "SkillAttackRange":0.5, "UpgradeGoldCoinCost":500, "UpgradeHealthGrowthValue":200, "RecoverHealthPerSec":50, "ReviveTime":10.0, "CanAttackAir":true, "CanAttackGround":true, "Description": "Your hero will restore a certain amount of health every second and will be revived after a certain period of time after death. The health value of your hero will not exceed its maximum health value, you can spend gold coins to increase the maximum health value of your hero.","SkillDescription": "Your hero has a skill called 'Fire of Rage' that requires you to actively release it. When you release this skill, the hero will ignite a raging fire at its feet. Any ground units (including your own knights) near this fire will be damaged, and the fire will continue to burn at this location for a period of time. This skill has no cooldown, but your hero will lose a lot of health when released. When this skill causes the death of your own knights, there is a certain probability that 'Friendly Fire Compensation' will be triggered, and you will receive a certain amount of gold coins as compensation."}
\end{lstlisting}

\begin{lstlisting}[caption=Configuration Table of Knight Reinforcements, language=json]
{"Number":2, "ExistTime":10.0, "Description": "You can also directly send knight reinforcements to the battlefield. This squad of knights will exist on the battlefield for a certain period of time, and then they will disappear. After they disappear, you can send them again."}
\end{lstlisting}

\begin{lstlisting}[caption=Configuration Table of Enemies,language=json]
{"Enemies":[
  {"Type":0,"FilePath":"Prefabs/enemy_soldier","Health":500,"MovementSpeed":0.5,"AttackSpeed":0.8, "AttackDamage":100,"AttackExtraDamage":50, "Name":"Orc Warrior","MovementType":"Ground", "Description": "Orc warriors move on the ground, they are heavily armored, so they have high health value."},
  {"Type":1,"FilePath":"Prefabs/enemy_wizard","Health":500,"MovementSpeed":0.5,"AttackSpeed":0.8, "AttackDamage":10,"AttackExtraDamage":10, "Name":"Orc Sorcerer","MovementType":"Ground","Description": "Orc Sorcerers move on the ground, they have powerful perception and magic abilities, and can freeze any defense tower that wants to attack it."},
  {"Type":2,"FilePath":"Prefabs/enemy_air","Health":550,"MovementSpeed":0.8,"AttackSpeed":0.8, "AttackDamage":10, "AttackExtraDamage":0,"Name":"Demon Bat","MovementType":"Flying","Description": "Demon Bats move across the sky, they don't attack, they just fly towards their destination quickly."},
  {"Type":3,"FilePath":"Prefabs/enemy_clown","Health":400,"MovementSpeed":0.5,"AttackSpeed":1.0, "AttackDamage":300,"AttackExtraDamage":200, "Name":"Clown","MovementType":"Ground","Description": "The clowns have skilled combat skills, so they have high attack power and attack speed, which can cause heavy damage to your knights and hero."},
  {"Type":4,"FilePath":"Prefabs/enemy_greeney","Health":400,"MovementSpeed":0.65,"AttackSpeed":0.8, "AttackDamage":120,"AttackExtraDamage":30, "Name":"Troll","MovementType":"Ground","Description": "Trolls have high movement speed and can easily pass through your defenses if you are not careful."},
  {"Type":5,"FilePath":"Prefabs/enemy_zombie","Health":500,"MovementSpeed":0.5,"AttackSpeed":1.0, "AttackDamage":80,"AttackExtraDamage":20, "Name":"Zombie","MovementType":"Ground","Description": "Zombies always appear in groups, and numbers are their advantage."},
  {"Type":6,"FilePath":"Prefabs/enemy_bonesoldier","Health":600,"MovementSpeed":0.4,"AttackSpeed":0.8, "AttackDamage":150,"AttackExtraDamage":20, "Name":"Bone Soldier","MovementType":"Ground","Description": "Bone Soldiers move slower but have higher attack power."},
  {"Type":7,"FilePath":"Prefabs/enemy_bonechanter","Health":1000,"MovementSpeed":0.5,"AttackSpeed":1.2, "AttackDamage":30,"AttackExtraDamage":20, "Name":"Bone Chanter","MovementType":"Ground","Description": "Bone Chanters can not only freeze defense towers, but also has a high health value."},
  {"Type":8,"FilePath":"Prefabs/enemy_piratesailor","Health":800,"MovementSpeed":0.5,"AttackSpeed":0.8, "AttackDamage":100,"AttackExtraDamage":50, "Name":"Pirate Sailor","MovementType":"Ground","Description": "Pirate Sailors have high health."},
  {"Type":9,"FilePath":"Prefabs/enemy_piratecaptain","Health":300,"MovementSpeed":0.5,"AttackSpeed":1.2, "AttackDamage":100,"AttackExtraDamage":500, "Name":"Pirate Captain","MovementType":"Ground","Description": "Pirate Captains have low health but have the potential to deal very high amounts of damage."},
  {"Type":10,"FilePath":"Prefabs/enemy_duckman","Health":400,"MovementSpeed":2.0,"AttackSpeed":0.8, "AttackDamage":30,"AttackExtraDamage":20, "Name":"Duckman","MovementType":"Ground","Description": "Duckmen have extremely high movement speed, but are vulnerable to attack."},
  {"Type":11,"FilePath":"Prefabs/enemy_outlaw","Health":400,"MovementSpeed":0.35,"AttackSpeed":0.9, "AttackDamage":80,"AttackExtraDamage":20, "Name":"Outlaw","MovementType":"Ground","Description": "The Outlaws are a group of robbers who have not received professional military training and are inferior in quality in all aspects."},
  {"Type":12,"FilePath":"Prefabs/enemy_hillking","Health":100000,"MovementSpeed":0.2,"AttackSpeed":0.8, "AttackDamage":30000,"AttackExtraDamage":20, "Name":"Hill King","MovementType":"Ground","Description": "Hill Kings cannot be defeated, and if you see one, your best option might be to run away."},
  {"Type":13,"FilePath":"Prefabs/enemy_trexrider1","Health":1000,"MovementSpeed":0.7,"AttackSpeed":0.7, "AttackDamage":300,"AttackExtraDamage":100, "Name":"T-Rex Rider","MovementType":"Ground","Description": "T-Rex Riders are elite warriors with high qualities in all aspects."},
  {"Type":14,"FilePath":"Prefabs/enemy_piratecommander","Health":1100,"MovementSpeed":0.7,"AttackSpeed":1.2, "AttackDamage":30,"AttackExtraDamage":20, "Name":"Pirate Commander","MovementType":"Ground","Description": "Pirate Commanders have high health and high movement speed."}
]}
\end{lstlisting}

\section{Environment Interface Additional Details}
\label{appendix:env_interface_details}
\begin{figure}
    \centering
    \includegraphics[width=1.0\linewidth]{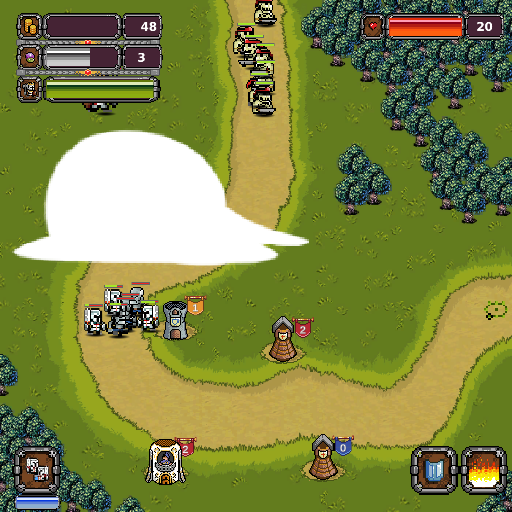}
    \caption{Pixel-based Observation of the TowerMind Environment.}
    \label{fig:obs}
\end{figure}

The observation space in TowerMind has three modalities: pixel-based, textual, and structured game-state observations: (1) Pixel-based Observation, it is a $512 \times 512$ RGB image representing the raw game screen of current step, as shown in Figure~\ref{fig:obs}; (2) Textual Observation, it is a JSON-formatted text that contains the game state at the current step, as shown in Listing~\ref{lst:text_obs}; (3) Structured Game-State Observation, it is obtained by extracting the numerical values from the textual observation's JSON and flattening them into a one-dimensional array of length 759, as shown in Table~1.

\begin{figure}
    \centering
    \includegraphics[width=1.0\linewidth]{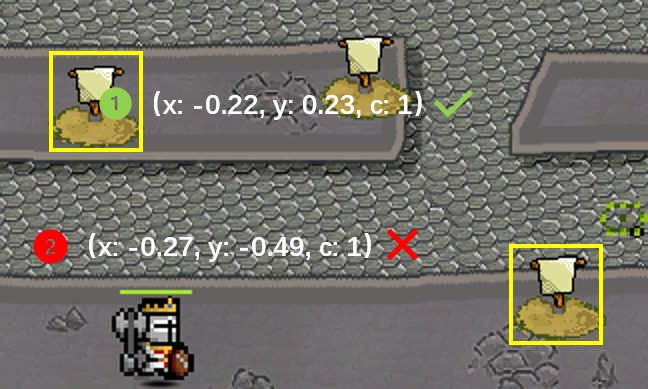}
    \caption{Each predefined tower point is associated with a bounding box of dimensions 0.5 × 0.5, as indicated by the yellow squares in the figure. For an action involving tower construction or related operations to be considered valid, the coordinates provided must lie within one of these bounding boxes. Otherwise, the action is treated as invalid and will not be executed. In the example depicted, both Action 1 and Action 2 aim to construct a Magician Tower, the circular numbered icons indicate the specific execution locations of the two actions. Action 1 specifies a location that falls within a valid bounding box and is thus valid and executable. Action 2, however, does not intersect with any valid bounding region and is consequently invalid and ignored.}
    \label{fig:actions_example}
\end{figure}

Actions are categorized as valid or invalid based on whether they conform to the game rules and the current state. A valid action is one that can be executed under the given constraints, whereas an invalid action violates these constraints and is therefore ignored. The validity of an action is influenced by several factors, including its target coordinates, the amount of available gold, and whether the hero unit is alive. For example, Figure~\ref{fig:actions_example} illustrates how the action's coordinates affect its validity.

The \texttt{Agent\_Last\_Action\_Info} field in the Textual Observation stores information about the agent’s action at the current step, including the action’s coordinates, type, execution status, and an associated error code if applicable. Different error codes indicate different reasons why an action may be considered invalid. A complete mapping of error codes and their corresponding meanings is provided in Listing~\ref{lst:error_code}.

\begin{lstlisting}[caption=Textual Observation of the TowerMind Environment.,label=lst:text_obs,language=json]
{
"Map_Center": {"X": 0.0, "Y": 0.0}, "Map_Side_Length": 6.0, "Map_Left_Boundary": -3.0, "Map_Right_Boundary": 3.0, "Map_Upper_Boundary": 3.0, "Map_Lower_Boundary": -3.0, "Tower_Points_Bounding_Box_Width_Height": 0.5, "Level_Gold_Coins_Collection_Count": 1, "Level_Friendly_Fire_Compensation_Count": 0, "Level_Maximum_Gold_Coins": 3000, "Level_Initial_Health": 20, "Level_Total_Waves_Number": 5, "Level_Inter_Wave_Interval": 6.0, "Level_Selling_Tower_Refund_Rate": 1.0, "Level_Gold_Coins_Refresh_Interval": 2, "Level_Gold_Coins_Retention_Time": 15, "Level_Gold_Coins_Minimum_Pickup_Amount": 100, "Level_Gold_Coins_Maximum_Pickup_Amount": 130, "Level_Enemy_Movement_Paths": [[{"X": -3.06, "Y": 0.28}, {"X": 3.09, "Y": 0.23}]], "Level_Enemy_Destination": {"X": 3.09, "Y": 0.23}, "Level_Current_Step": 192, "Level_Current_Time": 3.84, "Level_Current_Wave": 1, "Level_Current_Wave_Enemies": [0, 0, 0, 8, 0, 0, 0, 0, 0, 0, 0, 0, 0, 0, 0], "Level_Current_Wave_Countdown": "3", "Level_Current_Gold_Coins": 414, "Level_Current_Health": 20, "Level_Remaining_Waves": 5, "Level_Fog_Of_War_Position": {"X": -0.063, "Y": 1.424}, "Level_Knight_Reinforcements_Countdown": 0, "Level_Hero_Realtime_Status": {"Hero_Revive_Countdwon": 10, "Is_Hero_Dead": False, "Hero_Position": {"X": -0.985, "Y": -0.252}, "Hero_Current_Health": 1600}, "Level_Hero_Fire_Of_Rage_Positions": [], "Level_Towers_Realtime_Status": [{"Position": {"X": 1.68, "Y": 0.99}, "Tower_Name": "Knight Tower", "Is_Bulit": True, "Is_Frozen": False, "Knights_Assembly_Position": {"X": 1.65, "Y": 0.34}}, {"Position": {"X": -1.52, "Y": 0.9}, "Tower_Name": "Waiting to be Built", "Is_Bulit": False, "Is_Frozen": False, "Knights_Assembly_Position": {"X": -1.57, "Y": 0.21}}], "Level_Enemies_Realtime_Status": [], "Level_Knights_Realtime_Status": [{"Position": {"X": 0.162, "Y": 0.127}, "Name": "Knight", "Current_Health": 600}, {"Position": {"X": 1.621, "Y": 0.582}, "Name": "Knight", "Current_Health": 600}], "Level_Dropped_Gold_Coins
_Realtime_Status": {"Position": {"X": -1.613, "Y": -1.222}, "RemainingLifetime": 14}, "Agent_Last_Action_Info": {"Position": {"X": -1.613, "Y": -1.222}, "Action_Index": 9, "Is_Success": True, "Error_Code": 0}
}
\end{lstlisting}


\begin{table*}[ht]
\centering
  
  \label{tab:state_obs}
  \centering
  \fontsize{9pt}{9pt}\selectfont
  \begin{tabular}{llr}
    \toprule
    \textbf{Index}&\multicolumn{1}{c}{\textbf{Field}} & \multicolumn{1}{c}{\textbf{Dimension}}  \\
    \midrule
    1     & \texttt{Map\_Center} &2   \\
   \midrule
    2&\texttt{Map\_Side\_Length} &1   \\
    \midrule
    3&\texttt{Map\_Left\_Boundary} &1   \\
    \midrule
    4&\texttt{Map\_Right\_Boundary} &1   \\
    \midrule
    5&\texttt{Map\_Upper\_Boundary} &1   \\
    \midrule
    6&\texttt{Map\_Lower\_Boundary} &1   \\
    \midrule
    7&\texttt{Tower\_Points\_Bounding\_Box\_Width\_Height} &1   \\
    \midrule
    8&\texttt{Level\_Gold\_Coins\_Collection\_Count} &1   \\
    \midrule
    9&\texttt{Level\_Friendly\_Fire\_Compensation\_Count} &1   \\
    \midrule
    10&\texttt{Level\_Maximum\_Gold\_Coins} &1   \\
    \midrule
    11&\texttt{Level\_Initial\_Health} &1   \\
    \midrule
    12&\texttt{Level\_Total\_Waves\_Number} &1   \\
    \midrule
    13&\texttt{Level\_Inter\_Wave\_Interval} &1   \\
    \midrule
    15&\texttt{Level\_Selling\_Tower\_Refund\_Rate} &1   \\
    \midrule
    16&\texttt{Level\_Gold\_Coins\_Refresh\_Interval} &1   \\
    \midrule
    17&\texttt{Level\_Gold\_Coins\_Minimum\_Pickup\_Amount} &1   \\
    \midrule
    18&\texttt{Level\_Gold\_Coins\_Maximum\_Pickup\_Amount} &1   \\
    \midrule
    19&\texttt{Level\_Enemy\_Destination} &2   \\
    \midrule
    20&\texttt{Level\_Current\_Step} &1   \\
    \midrule
    21&\texttt{Level\_Current\_Time} &1   \\
    \midrule
    22&\texttt{Level\_Current\_Wave} &1   \\
    \midrule
    23&\texttt{Level\_Current\_Wave\_Countdown} &1   \\
    \midrule
    24&\texttt{Level\_Current\_Gold\_Coins} &1   \\
    \midrule
    25&\texttt{Level\_Current\_Health} &1   \\
    \midrule
    26&\texttt{Level\_Remaining\_Waves} &1   \\
    \midrule
    27&\texttt{Level\_Fog\_Of\_War\_Position} &2   \\
    \midrule
    28&\texttt{Level\_Knight\_Reinforcements\_Countdown} &1   \\
    \midrule
    29&\texttt{Level\_Hero\_Realtime\_Status} &5   \\
    \midrule
    30&\texttt{Level\_Dropped\_Gold\_Coins\_Realtime\_Status} &3   \\
    \midrule
    31&\texttt{Agent\_Last\_Action\_Info} &5   \\
    \midrule
    32&\texttt{Level\_Enemy\_Movement\_Paths} &5 * 20 * 2   \\
    \midrule
    33&\texttt{Level\_Current\_Wave\_Enemies} &25   \\
    \midrule
    34&\texttt{Level\_Hero\_Fire\_Of\_Rage\_Positions} &10 * 2   \\
    \midrule
    35&\texttt{Level\_Towers\_Realtime\_Status} &15 * 8   \\
    \midrule
    36&\texttt{Level\_Enemies\_Realtime\_Status} &50 * 4   \\
    \midrule
    37&\texttt{Level\_Knights\_Realtime\_Status} &50 * 3   \\
    \midrule
     &Total &759   \\
    \bottomrule
  \end{tabular}
\caption{Structured Game-State Observation of the TowerMind Environment.}
\end{table*}

\begin{lstlisting}[caption=Error Code Details.,label=lst:error_code,language=json]
0 - no error

1 - build a tower where there is already a tower

2 - build a tower but don't have enough gold coins

3 - upgrade a non-existent tower

4 - upgrade a tower but don't have enough gold coins

5 - sell a non-existent tower

6 - failure to provide valid coordinates for building, upgrading, selling a tower or showing the attack range of a tower

7 - failed to provide the valid coordinates for changing the knights assembly location of a Knight Tower

8 - deploy Knight Reinforcements that are on cooldown

9 - try to manipulate a dead hero

10 - increase your hero's maximum health but don't have enough gold coins

11 - show the attack range of a non-existent tower

\end{lstlisting}

\section{Additional Details of the Benchmark Levels}
\label{appendix:benchmark_levels}

Table~2 summarizes their level difficulty $D(l)$ and the contributing components, calculated using the level difficulty quantitative metric system introduced in main body of the paper: $D(l) = d_{rd}(l) + d_{t}(l) + d_{e}(l) + d_{re}(l)$. When computing the road component $d_{rd}(l)$, we set $R_{\text{max}} = 5$, the design-time maximum number of roads on any map. For the tower component $d_{t}(l)$, we use $T_{\text{max}} = 15$, the design-time maximum number of tower points on any map. For the enemy component $d_{e}(l)$, we set $N_{\text{max}} = 25$, representing the design-time maximum number of enemies per wave in this environment. Finally, when computing the resource component $d_{re}(l)$, we use $I_{\text{min}} = 120$ and $G_{\text{min}} = 40$, which denote the design-time minimum initial gold and the minimum gold drop per enemy, respectively.

\begin{table*}[ht]
\centering
\fontsize{9pt}{9pt}\selectfont
 \resizebox{\textwidth}{!}{ 
\begin{threeparttable}
  
  \label{tab:benchmark_levels_detail}
  \centering
  \begin{tabular}{ccccccccc}
    \toprule
    \textbf{Level}&\multicolumn{1}{c}{\textbf{\#Roads}} & \multicolumn{1}{c}{\textbf{\#Tower Points}}& \multicolumn{1}{c}{\textbf{\#Enemy Types}} & \multicolumn{1}{c}{\textbf{\#Enemies per Wave}}& \multicolumn{1}{c}{\textbf{Initial Gold Coins}}& \multicolumn{1}{c}{\textbf{Gold Coins Drop Amount}}&\multicolumn{1}{c}{\textbf{Tower Sell-back Ratio}}&\multicolumn{1}{c}{\textbf{Level Difficulty}} \\
    \midrule
    Lv1&1 &4&14&20.8&500&100&100\%   &2.45   \\
    \midrule
    Lv2&1 &5&13&9.2&120&40&0\%   &2.77\\
    \midrule
    Lv3&3 &12&14&12.0&500&60&10\%   &3.42\\
    \midrule
    Lv4&3 &12&14&17.0&500&70&20\%&3.55   \\
    \midrule
    
    Lv5 &4&13&11&16.4&500&50&0\%&3.74   \\
    \bottomrule
  \end{tabular}
  
  \begin{tablenotes}
\item[*]\#Roads is the number of roads in this level; \#Tower Points is the number of tower points in this level; \#Enemy Types is the number of enemy types in this level; \#Enemies per Wave is the average number of enemies per wave in this level; Initial Gold Coins is the initial gold coins in this level; Gold Coins Drop Amount is the amount of gold coins dropped each time in this level;
\end{tablenotes}
  \end{threeparttable}
  
  }
  \caption{The information of the benchmark levels' difficulty and their contributing components.}
\end{table*}

\section{Additional Details of Environment Customizability}
\label{appendix:env_cus}
\begin{figure}
    \centering
    \includegraphics[width=1.0\linewidth]{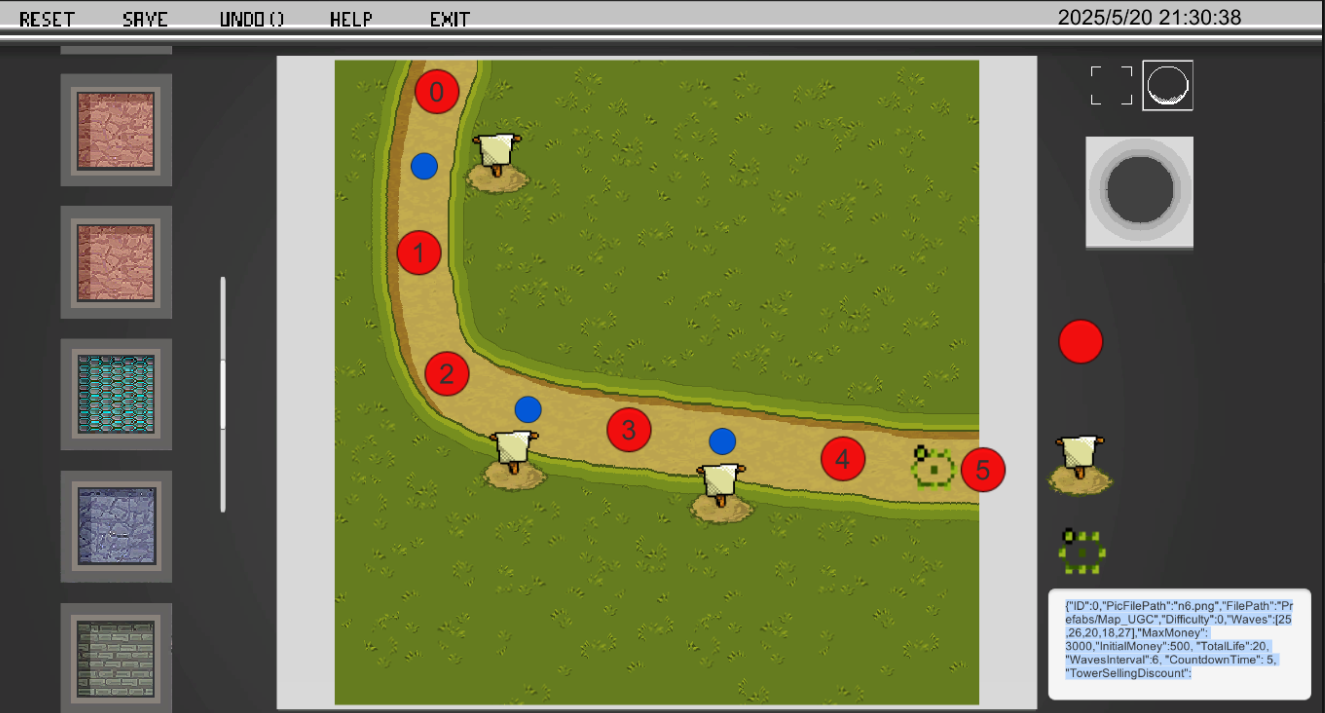}
    \caption{This is a screenshot of the TowerMind level editor. On the left is the background pattern selector for the map. The central area is the map editing workspace, where red dots represent waypoints along the road, and blue dots indicate the default summoning points for knights at corresponding tower points. On the right, the upper section contains brush settings for map editing, while the lower section provides a drag-and-drop interface for placing game elements.}
    \label{fig:levels_editor}
\end{figure}

The TowerMind level editor is a user-friendly, GUI-based application (as illustrated in Figure~\ref{fig:levels_editor}). It allows users to design custom maps by drawing various road shapes using a brush tool. Core gameplay elements such as road waypoints and tower points can be added to the map via a drag-and-drop interface. Once the design is complete, users can export the map as an image along with a corresponding JSON file, which can then be imported into the TowerMind environment for use.

All configuration files for the TowerMind environment are located in the \texttt{StreamingAssets/Config} directory (the exact path may vary by platform; see the user documentation for details). All configuration files are in JSON format and can be freely modified as needed.

Among all configuration files, \texttt{EnvConfig} manages auxiliary features such as enabling action space discretization preview and human player gameplay recording. For more details, please refer to the user documentation.

\section{Zero-shot Prompt}
\label{appendix:prompt}
The prompt consists of four components: (1) an explanation of the game's overall objective and core rules in natural language; (2) an explanation of the action space in natural language; (3) the raw configuration tables for towers, knight units, the hero unit, knight reinforcements, and enemies in JSON format, listing their numerical attributes; (4) the observation–action pair history in JSON format, with a default length of 3, showing the agent’s recent interactions with the environment.

This is a complete prompt example, excluding the configuration details for towers, knight units, the hero unit, knight reinforcements, and enemies, which are provided in the Appendix~\ref{appendix:game_mechanics_details}.

\begin{tcolorbox}[title=An Prompt Example]
You are an AI agent playing a video game, you need to build different types of defense towers at different locations on the map to prevent enemies from reaching their destination.
\vspace{1em}

Common rules:

- You need to spend gold coins to build towers, upgrade towers and increase your hero's maximum health. Gold coins will continue to drop at random locations on the map. You can send your knights or hero to pick up gold coins, and the gold coins will be picked up automatically when your knights or hers are near the gold coins.

- If the number of gold coins you hold exceeds the maximum value, the excess will be discarded.

- You will be given a certain amount of health at the beginning of each level. Every time an enemy reaches its destination, you will lose a point of health. When your health reaches 0, the game ends and the mission fails. Try your best to avoid losing any health points.

- Enemies appear in waves, and each level has a different number of enemy waves. There is a certain amount of time between enemy waves. If your health is still greater than 0 after you have resisted all waves of enemy attacks, the mission is successful.

- There are several paths for the enemies, and each enemy will randomly choose one.

- The battlefield of this game is a square area, the details have been included in the level state part blow.

- Between each path point, enemies will only move in a straight line.

- The Fog Of War in the battlefield is an irregular cloud-shaped area that can obscure any element in the game. Its approximate dimensions are 3.5 wide and 1.7 tall. The obscured towers, knights and heroes will no longer attack the enemy, but if the Fog Of War obscures the Fire Of Rage released by the hero, it will lose its obstruction ability during this time.

\vspace{1em}

The following are the actions you can take. And for each action, you also need to provide a horizontal and a vertical coordinate between -3.0 and 3.0.

0 - Build an Archer Tower at the coordinates you specify,

1 - Build an Magician Tower at the coordinates you specify,

2 - Build an Knight Tower at the coordinates you specify,

3 - Upgrade a tower at the coordinates you specify,

4 - Sell a tower at the coordinates you specify,

5 - Show the attack range of a tower at the coordinates you specify,

6 - Noop: do nothing,

7 - Change the knights assembly location of a Knight Tower to the coordinates you specify,

8 - Deploy Knight Reinforcements to the coordinates you specify,

9 - Dispatch your hero to the coordinates you specify,

10 - Your hero casts 'Fire of Rage' at your hero's coordinates,

11 - Spend gold coins to increase your hero's maximum health.

\vspace{1em}

Action Tips:

- Building a tower, upgrading a tower or increasing your hero's maximum health requires you to have enough gold coins, otherwise it will be an invalid action.

- Action 0, 1, 2, 3, 4, 5 are only valid if the coordinates you specify are within the bounding box of the tower point. The bounding box of the tower point is a square with its coordinate as the center and a side length of 0.5.

- Action 7 is only valid if the coordinates you specify is within the attack range of a Knight Tower.

- Action 8 will be invalid during the Knight Reinforcements cooldown.

- Action 9 means that your hero starts moving to the coordinates you specify, not a direct teleportation. If you set a new target coordinate during its movement, it will start moving to the new target coordinates.

- Actions 9, 10, 11 are invalid if your hero dies.

- If a tower point already has a tower, you should not build a tower at this tower point, which will result in an invalid action.

- You should provide your action in json format, only three elements in this json structure: "X" is a floating point number representing the horizontal coordinate of the action you want to perform; "Y" is a floating point number representing the vertical coordinate of the action you want to perform; "Action" is an integer representing the index of action you want to perform. 

- Action 4 will return the funds spent on its construction and upgrade, but it may not be fully refunded, it depends on the '\texttt{Level\_Selling\_Tower\_Refund\_Rate}' value.

\vspace{1em}

The following are the actions error code list, If you performed an invalid action, you can find out why here:

0 - no error

1 - build a tower where there is already a tower

2 - build a tower but don't have enough gold coins

3 - upgrade a non-existent tower

4 - upgrade a tower but don't have enough gold coins

5 - sell a non-existent tower

6 - failure to provide valid coordinates for building, upgrading, selling a tower or showing the attack range of a tower

7 - failed to provide the valid coordinates for changing the knights assembly location of a Knight Tower

8 - deploy Knight Reinforcements that are on cooldown

9 - try to manipulate a dead hero

10 - increase your hero's maximum health but don't have enough gold coins

11 - show the attack range of a non-existent tower

\vspace{1em}

The following is the configuration table of each component of the game, organized in Json format:

- Towers Configuration:
\begin{lstlisting}[language=json]
{...}
\end{lstlisting}

- Knight Configuration:
\begin{lstlisting}[language=json]
{...}
\end{lstlisting}

- Hero Configuration:
\begin{lstlisting}[language=json]
{...}
\end{lstlisting}

- Knight Reinforcements Configuration:
\begin{lstlisting}[language=json]
{...}
\end{lstlisting}

- Enemies Configuration:
\begin{lstlisting}[language=json]
{...}
\end{lstlisting}

\vspace{1em}

Configuration Table Tips:

- The attack range of the towers, hero, hero's skill and knights is circular, the positions of the circle centers are their position, and the attack range described above is the diameter. When enemies enter this range they will attack.

- The final attack value of the towers, hero, hero's skill, knights and enemies is equal to AttackDamage plus a random value in the range of 0 to AttackExtraDamage.

- The unit of time in this tower defense game is seconds.

- The unit of range or space in this tower defense game is a virtual unified unit. It can be used directly for calculation during reasoning without conversion.

- The AttackSpeed of the towers, hero, knights and enemies refers to the time interval between attacks. For the Knight Tower, it refers to the time interval between summoning knights.

- Upgrading will increase the attack power of the Archer Tower and the Magician Tower, as well as the attack value and movement speed of the knights summoned by the Knight Tower.

\vspace{1em}

The following is the information about this level, organized in Json format:
\begin{lstlisting}[language=json]
{
    "Map_Center": {"X": 0.0, "Y": 0.0}, 
    "Map_Side_Length": 6.0, 
    "Map_Left_Boundary": -3.0, 
    "Map_Right_Boundary": 3.0, 
    "Map_Upper_Boundary": 3.0, 
    "Map_Lower_Boundary": -3.0,
    "Tower_Points_Bounding_Box_
    Width_Height": 0.5,
    "Level_Maximum_Gold_Coins": 3000, 
    "Level_Initial_Health": 20, 
    "Level_Total_Waves_Number": 5, 
    "Level_Inter_Wave_Interval": 6.0, 
    "Level_Selling_Tower_Refund_
    Rate": 1.0,
    "Level_Gold_Coins_Refresh_
    Interval": 2,
    "Level_Gold_Coins_Retention_
    Time": 15,
    "Level_Gold_Coins_Maximum_
    Pickup_Amount": 130,
    "Level_Gold_Coins_Minimum_
    Pickup_Amount": 100,
    "Level_Enemy_Movement_Paths": [[{"X": -3.06, "Y": 0.28}, {"X": 3.09, "Y": 0.23}]], 
    "Level_Enemy_Destination": {"X": 3.09, "Y": 0.23}
}
\end{lstlisting}

\vspace{1em}

The following is the history of the past few steps, organized in Json format:
\begin{lstlisting}[language=json]
{"state": "{"Level_Gold_Coins_
Collection_Count": 1, "Level_Friendly_Fire_
Compensation_Count": 0, "Level_Current_Step": 176, "Level_Current_Time": 3.52, "Level_Current_Wave": 1, "Level_Current_Wave_Enemies": [0, 0, 0, 8, 0, 0, 0, 0, 0, 0, 0, 0, 0, 0, 0], "Level_Current_Wave_Countdown": "3", "Level_Current_Gold_Coins": 254, "Level_Current_Health": 20, "Level_Remaining_Waves": 5, "Level_Fog_Of_War_Position": {"X": -0.049, "Y": 1.486}, "Level_Knight_Reinforcements_
Countdown": 0, "Level_Hero_Realtime_Status": {"Hero_Revive_Countdwon": 10, "Is_Hero_Dead": False, "Hero_Position": {"X": -0.597, "Y": -0.183}, "Hero_Current_Health": 1600}, "Level_Hero_Fire_Of_Rage_
Positions": [], "Level_Towers_Realtime_Status": [{"Position": {"X": 1.68, "Y": 0.99}, "Tower_Name": "Archer Tower", "Is_Bulit": True, "Is_Frozen": False, "Knights_Assembly_Position": {"X": 1.65, "Y": 0.34}}, {"Position": {"X": -1.52, "Y": 0.9}, "Tower_Name": "Archer Tower", "Is_Bulit": True, "Is_Frozen": False, "Knights_Assembly_Position": {"X": -1.57, "Y": 0.21}}], "Level_Enemies_Realtime_Status": [], "Level_Knights_Realtime_Status": [], "Level_Dropped_Gold_Coins_
Realtime_Status": {"Position": {"X": 2.191, "Y": -2.272}, "RemainingLifetime": 14}, "Agent_Last_Action_Info": {"Position": {"X": 2.191, "Y": -2.272}, "Action_Index": 9, "Is_Success": True, "Error_Code": 0}}", "action": array([ 2.191, -2.272,  9.   ])},
{"state": "{"Level_Gold_Coins_Collection_
Count": 1, "Level_Friendly_Fire_
Compensation_Count": 0, "Level_Current_Step": 192, "Level_Current_Time": 3.84, "Level_Current_Wave": 1, "Level_Current_Wave_Enemies": [0, 0, 0, 8, 0, 0, 0, 0, 0, 0, 0, 0, 0, 0, 0], "Level_Current_Wave_Countdown": "3", "Level_Current_Gold_Coins": 254, "Level_Current_Health": 20, "Level_Remaining_Waves": 5, "Level_Fog_Of_War_Position": {"X": -0.063, "Y": 1.424}, "Level_Knight_Reinforcements_
Countdown": 0, "Level_Hero_Realtime_Status": {"Hero_Revive_Countdwon": 10, "Is_Hero_Dead": False, "Hero_Position": {"X": -0.367, "Y": -0.355}, "Hero_Current_Health": 1600}, "Level_Hero_Fire_Of_Rage_
Positions": [], "Level_Towers_
Realtime_Status": [{"Position": {"X": 1.68, "Y": 0.99}, "Tower_Name": "Archer Tower", "Is_Bulit": True, "Is_Frozen": False, "Knights_Assembly_Position": {"X": 1.65, "Y": 0.34}}, {"Position": {"X": -1.52, "Y": 0.9}, "Tower_Name": "Archer Tower", "Is_Bulit": True, "Is_Frozen": False, "Knights_Assembly_Position": {"X": -1.57, "Y": 0.21}}], "Level_Enemies_Realtime_Status": [], "Level_Knights_Realtime_Status": [], "Level_Dropped_Gold_Coins_
Realtime_Status": {"Position": {"X": 2.191, "Y": -2.272}, "RemainingLifetime": 14}, "Agent_Last_Action_Info": {"Position": {"X": 2.191, "Y": -2.272}, "Action_Index": 9, "Is_Success": True, "Error_Code": 0}}", "action": array([ 2.191, -2.272,  9.   ])},
{"state": "{"Level_Gold_Coins_Collection_
Count": 1, "Level_Friendly_Fire_
Compensation_Count": 0, "Level_Current_Step": 208, "Level_Current_Time": 4.16, "Level_Current_Wave": 1, "Level_Current_Wave_Enemies": [0, 0, 0, 8, 0, 0, 0, 0, 0, 0, 0, 0, 0, 0, 0], "Level_Current_Wave_Countdown": "2", "Level_Current_Gold_Coins": 254, "Level_Current_Health": 20, "Level_Remaining_Waves": 5, "Level_Fog_Of_War_Position": {"X": -0.078, "Y": 1.361}, "Level_Knight_Reinforcements_
Countdown": 0, "Level_Hero_Realtime_Status": {"Hero_Revive_Countdwon": 10, "Is_Hero_Dead": False, "Hero_Position": {"X": -0.137, "Y": -0.528}, "Hero_Current_Health": 1600}, "Level_Hero_Fire_Of_Rage_
Positions": [], "Level_Towers_Realtime_Status": [{"Position": {"X": 1.68, "Y": 0.99}, "Tower_Name": "Archer Tower", "Is_Bulit": True, "Is_Frozen": False, "Knights_Assembly_Position": {"X": 1.65, "Y": 0.34}}, {"Position": {"X": -1.52, "Y": 0.9}, "Tower_Name": "Archer Tower", "Is_Bulit": True, "Is_Frozen": False, "Knights_Assembly_Position": {"X": -1.57, "Y": 0.21}}], "Level_Enemies_Realtime_Status": [], "Level_Knights_Realtime_Status": [], "Level_Dropped_Gold_Coins_
Realtime_Status": {"Position": {"X": 2.191, "Y": -2.272}, "RemainingLifetime": 13}, "Agent_Last_Action_Info": {"Position": {"X": 2.191, "Y": -2.272}, "Action_Index": 9, "Is_Success": True, "Error_Code": 0}}", "action": array([ 2.191, -2.272,  9.   ])}\end{lstlisting}

\vspace{1em}

The following is the current real-time game status of this step, organized in Json format:
\begin{lstlisting}[language=json]
{
    "Level_Gold_Coins_Collection_
    Count": 1,
    "Level_Friendly_Fire_
    Compensation_Count": 0, 
    "Level_Current_Step": 224, 
    "Level_Current_Time": 4.48, 
    "Level_Current_Wave": 1, 
    "Level_Current_Wave_Enemies": [0, 0, 0, 8, 0, 0, 0, 0, 0, 0, 0, 0, 0, 0, 0], 
    "Level_Current_Wave_Countdown": "2", 
    "Level_Current_Gold_Coins": 254, 
    "Level_Current_Health": 20, 
    "Level_Remaining_Waves": 5, 
    "Level_Fog_Of_War_Position": {"X": -0.092, "Y": 1.299},
    "Level_Knight_Reinforcements_
    Countdown": 0,
    "Level_Hero_Realtime_Status": {
        "Hero_Revive_Countdwon": 10, 
        "Is_Hero_Dead": False, 
        "Hero_Position": {"X": 0.094, "Y": -0.701},
        "Hero_Current_Health": 1600
        }, 
    "Level_Hero_Fire_Of_Rage_
    Positions": [],
    "Level_Towers_Realtime_Status": [
        {"Position": {"X": 1.68, "Y": 0.99}, 
        "Tower_Name": "Archer Tower", 
        "Is_Bulit": True, 
        "Is_Frozen": False, 
        "Knights_Assembly_Position": {"X": 1.65, "Y": 0.34}},
        {"Position": {"X": -1.52, "Y": 0.9}, 
        "Tower_Name": "Archer Tower", 
        "Is_Bulit": True, 
        "Is_Frozen": False, 
        "Knights_Assembly_Position": {"X": -1.57, "Y": 0.21}}
        ], 
    "Level_Enemies_Realtime_Status": [],
    "Level_Knights_Realtime_Status": [],
    "Level_Dropped_Gold_Coins_
    Realtime_Status": {
        "Position": {"X": 2.191, "Y": -2.272}, 
        "RemainingLifetime": 13
        }, 
    "Agent_Last_Action_Info": {
        "Position": {"X": 2.191, "Y": -2.272}, 
        "Action_Index": 9, 
        "Is_Success": True, 
        "Error_Code": 0
        }
}\end{lstlisting}

\vspace{1em}

Image observation provided.

\vspace{1em}

Now please tell me the action you want to perform in this step, in JSON format, containing a floating point X coordinate, a floating point Y coordinate and an integer action index. Your answer should not contain any other text, just provide this json.

\end{tcolorbox}

\begin{figure}[htbp]
  \centering
  \includegraphics[clip=false, width=\linewidth]{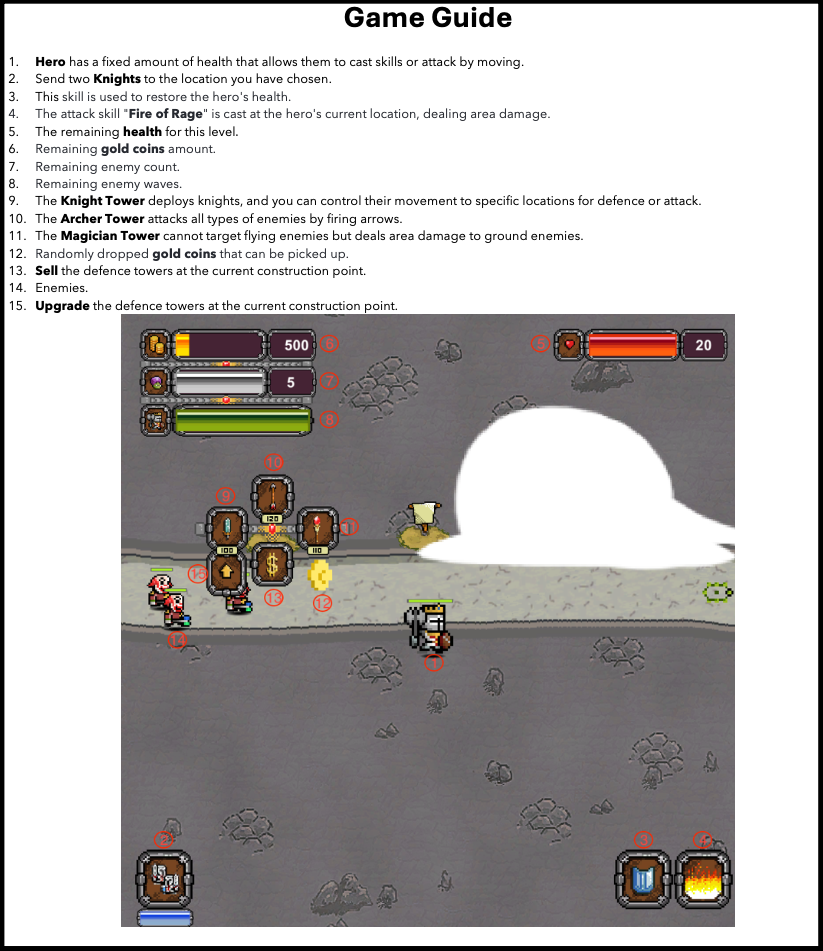}
  \caption{The game guide for human participants.}
  \label{fig:game_guide}
\end{figure}

\section{Human Expert Performance}
\label{appendix:human_expert}
Five participants with prior experience in tower defense games were recruited from a university environment. All participants expressed willingness and availability to engage in approximately 15 hours of study-related activities within a one-week period. Each participant signed an informed consent form, which detailed the study's objectives, methodology, potential risks and discomforts, expected benefits, financial considerations, confidentiality terms, and their rights throughout and following their involvement in the study. Each participant was compensated at a rate of £17.89 per hour for their time.

All five participants were provided with a game guide for the TowerMind environment (as shown in Figure~\ref{fig:game_guide}), along with five non-benchmark levels for training purposes. The participants conducted their training sessions on their personal laptops using a mouse and keyboard, and we verified that the TowerMind environment ran smoothly on all of their devices. Following 13 hours of training, each participant was evaluated on the five benchmark levels, with a unique random seed assigned to each. Their evaluation metrics were the same as those used for the LLMs: score and valid action rate. Their APM during evaluation ranged from 40 to 110. The evaluation results are summarized in Table~3.

\begin{table}[ht]
  
  \label{tab:human_expert_eval}
  \centering
  \fontsize{9pt}{9pt}\selectfont
  \begin{tabular}{lrr}
    \toprule
    \textbf{Level} & \textbf{Score} & \textbf{Valid Action Rate} \\
    \midrule
    LV 1 & $0.00 \pm 0.20$ & $0.99 \pm 0.01$ \\
    LV 2 & $-0.20 \pm 0.00$ & $0.98 \pm 0.03$ \\
    LV 3 & $-0.40 \pm 0.24$ & $0.97 \pm 0.01$ \\
    LV 4 & $-1.80 \pm 0.68$ & $0.94 \pm 0.00$ \\
    LV 5 & $-3.40 \pm 0.49$ & $0.93 \pm 0.01$ \\
    \midrule
    \textbf{Average} & $\mathbf{-1.16 \pm 0.40}$ & $\mathbf{0.96 \pm 0.01}$ \\
    \bottomrule
  \end{tabular}
  \caption{Average performance of five human experts across five benchmark levels. Scores range from –20 to 0, and valid action rates range from 0 to 1. Standard errors are included.}
\end{table}

\section{Additional Experimental Results on LLM Evaluation}
\label{appendix:additional_experimental_results}

All evaluations of LLMs were conducted on a single laptop equipped with a 13th Gen Intel(R) Core(TM) i7-13650HX CPU (2.60 GHz), 64 GB of RAM, and an NVIDIA GeForce RTX 4060 Laptop GPU with 8 GB of VRAM. The average evaluation time per episode in the language-only modality is approximately 7 minutes, while the vision-language modality requires 15 minutes per episode. Figures~4 and 5 show the score and valid action rate performance with standard errors, respectively. All values are normalized relative to the human expert baseline.

Given the level score or valid action rate of an LLM on level \( l \), \( s_l^{\text{(raw)}} \), we compute the normalized human-relative score or valid action rate \( s_l \) from the human baseline \( s_l^{\text{(human)}} \) and the minimum possible game score or valid action rate \( s_l^{\text{(min)}} \):
\[
s_l = \frac{ s_l^{\text{(raw)}} - s_l^{\text{(min)}}}{s_l^{\text{(human)}} - s_l^{\text{(min)}}}
\]

See the supplementary material (Section: Additional LLM Evaluation Results) for correlation analysis between score and valid action rate.

\begin{table*}[ht]
  
  \label{tab:reward-performance-with-se}
  \centering
  \fontsize{9pt}{9pt}\selectfont
  \begin{tabular}{l*{9}{c}c}
    \toprule
    \textbf{Model} &  \textbf{LV 1} & \textbf{LV 2} & \textbf{LV 3} & \textbf{LV 4} & \textbf{LV 5} & \textbf{Average} \\

    \midrule
    \multicolumn{7}{l}{\textbf{Language-Only}} \\
    GPT-4.1                     & $0.59 \pm0.17$ & $0.49 \pm0.06$ & $0.32 \pm0.04$ & $0.19 \pm0.12$ & $0.07 \pm0.04$ & $0.33 \pm0.10$ \\
    Gemini-1.5-Pro              & $0.52\pm0.14$  & $0.42 \pm0.14$ & $0.31 \pm0.01$ & $0.11 \pm0.13$ & $0.01 \pm0.03$ & $0.27 \pm0.09$ \\
    Claude 3.7 Sonnet          & $0.62 \pm0.07$ & $0.51 \pm0.07$ & $0.40 \pm0.06$ & $0.24 \pm0.19$ & $0.15 \pm0.10$ & $0.38 \pm0.07$ \\
    Llama 3.2 90B               & $0.42 \pm0.02 $& $0.32 \pm0.13 $& $0.19 \pm0.09 $& $0.12 \pm0.00 $& $0.00 \pm0.00 $& $0.21 \pm0.10 $ \\
    Llama 3.2 11B               & $0.17 \pm0.13 $& $0.09 \pm0.03 $& $0.00 \pm0.00 $& $0.00 \pm0.00 $& $0.00 \pm0.00 $& $0.05 \pm0.08 $ \\
    Qwen 2.5-VL 72B              & $0.47 \pm0.10 $& $0.36 \pm0.10 $& $0.21 \pm0.16 $& $0.00 \pm0.00 $& $0.00 \pm0.00 $& $0.21 \pm0.10 $ \\
    Qwen 2.5-VL 7B              & $0.00 \pm0.00 $& $0.00 \pm0.00 $& $0.00 \pm0.00 $& $0.00 \pm0.00 $& $0.00 \pm0.00 $& $0.00 \pm0.00 $ \\
    \midrule
    \multicolumn{7}{l}{\textbf{Vision-Language}} \\
    GPT-4.1              & $0.63 \pm0.19 $& $0.56 \pm0.10 $& $0.44 \pm0.12 $& $0.32 \pm0.20 $& $0.15 \pm0.09 $& $0.42 \pm0.08 $ \\
    Gemini-1.5-Pro       & $0.57 \pm0.19 $& $0.44 \pm0.12 $& $0.33 \pm0.13 $& $0.16 \pm0.06 $& $0.01 \pm0.00 $& $0.30 \pm0.09 $ \\
    Claude 3.7 Sonnet    & $0.67 \pm0.07 $& $0.58 \pm0.06 $& $0.45 \pm0.07 $& $0.20 \pm0.07 $& $0.16 \pm0.10 $& $0.41 \pm0.09 $ \\
    Llama 3.2 90B         & $0.30 \pm0.13 $& $0.05 \pm0.05 $& $0.00 \pm0.00 $& $0.00 \pm0.00 $& $0.00 \pm0.00 $& $0.07 \pm0.08 $ \\
    Llama 3.2 11B         & $0.04 \pm0.04 $& $0.00 \pm0.00 $& $0.00 \pm0.00 $& $0.00 \pm0.00 $& $0.00 \pm0.00 $& $0.01 \pm0.04 $ \\
    Qwen 2.5-VL 72B        & $0.54 \pm0.09 $& $0.39 \pm0.12 $& $0.20 \pm0.10 $& $0.12 \pm0.18 $& $0.05 \pm0.05 $& $0.26 \pm0.11 $ \\
    Qwen 2.5-VL 7B         & $0.05 \pm0.15 $& $0.00 \pm0.00 $& $0.00 \pm0.00 $& $0.00 \pm0.00 $& $0.00 \pm0.00 $& $0.01 \pm0.11 $ \\
    \midrule
    Random                     &$0.00 \pm 0.00$&$0.00 \pm 0.00$&$0.00 \pm 0.00$&$0.00 \pm 0.00$&$0.00 \pm 0.00$& $0.00 \pm 0.00 $                                               \\
    \bottomrule
  \end{tabular}
  \caption{The score performance of LLMs on the benchmark levels with standard error.}
\end{table*}

\begin{table*}[ht]
  
  \label{tab:rate-performance-with-se}
  \centering
  \fontsize{9pt}{9pt}\selectfont
  \begin{tabular}{l*{9}{c}c}
    \toprule
    \textbf{Model} &  \textbf{LV 1} & \textbf{LV 2} & \textbf{LV 3} & \textbf{LV 4} & \textbf{LV 5} & \textbf{Average} \\
    \midrule
    \multicolumn{7}{l}{\textbf{Language-Only}} \\
    GPT-4.1                     & $0.92 \pm0.01$ & $0.89 \pm0.02$ & $0.88 \pm0.02$ & $0.84 \pm0.02$ & $0.75 \pm0.02$ & $0.86 \pm0.02$ \\
    Gemini-1.5-Pro              & $0.91 \pm0.02$ & $0.90 \pm0.02$ & $0.89 \pm0.03$ & $0.83 \pm0.04$ & $0.82 \pm0.02$ & $0.87 \pm0.02$ \\
    Claude 3.7 Sonnet           & $0.90 \pm0.04$ & $0.87 \pm0.03$ & $0.85 \pm0.02$ & $0.85 \pm0.03$ & $0.79 \pm0.02$ & $0.85 \pm0.02$ \\
    Llama 3.2 90B              & $0.48 \pm0.02 $& $0.39 \pm0.02 $& $0.30 \pm0.02 $& $0.21 \pm0.03 $& $0.20 \pm0.03 $& $0.32 \pm0.04 $ \\
    Llama 3.2 11B               & $0.28 \pm0.02 $& $0.24 \pm0.03 $& $0.23 \pm0.03 $& $0.23 \pm0.03 $& $0.22 \pm0.06 $& $0.24 \pm0.02 $ \\
    Qwen 2.5-VL 72B              & $0.87 \pm0.04 $& $0.78 \pm0.08 $& $0.76 \pm0.10 $& $0.58 \pm0.08 $& $0.51 \pm0.08 $& $0.70 \pm0.05 $ \\
    Qwen 2.5-VL 7B               & $0.11 \pm0.01 $& $0.05 \pm0.00 $& $0.03 \pm0.10 $& $0.01 \pm0.01 $& $0.01 \pm0.01 $& $0.04 \pm0.01 $ \\
    
    \midrule
    \multicolumn{7}{l}{\textbf{Vision-Language}} \\
    GPT-4.1              & $0.86 \pm0.01 $& $0.81 \pm0.02 $& $0.75 \pm0.02 $& $0.68 \pm0.03 $& $0.66 \pm0.02 $& $0.75 \pm0.03 $ \\
    Gemini-1.5-Pro        & $0.85 \pm0.05 $& $0.81 \pm0.09 $& $0.80 \pm0.05 $& $0.73 \pm0.05 $& $0.67 \pm0.04 $& $0.77 \pm0.03 $ \\
    Claude 3.7 Sonnet    & $0.85 \pm0.02 $& $0.85 \pm0.01 $& $0.83 \pm0.01 $& $0.80 \pm0.02 $& $0.79 \pm0.02 $& $0.82 \pm0.02 $ \\
    Llama 3.2 90B         & $0.44 \pm0.06 $& $0.38 \pm0.07 $& $0.33 \pm0.08 $& $0.31 \pm0.04 $& $0.30 \pm0.09 $& $0.35 \pm0.03 $ \\
    Llama 3.2 11B        & $0.31 \pm0.02 $& $0.19 \pm0.12 $& $0.18 \pm0.04 $& $0.13 \pm0.04 $& $0.11 \pm0.03 $& $0.18 \pm0.02 $ \\
    Qwen 2.5-VL 72B        & $0.79 \pm0.04 $& $0.72 \pm0.06 $& $0.66 \pm0.07 $& $0.54 \pm0.02 $& $0.43 \pm0.04 $& $0.63 \pm0.04 $ \\
    Qwen 2.5-VL 7B        & $0.21 \pm0.02 $& $0.15 \pm0.01 $& $0.05 \pm0.02 $& $0.04 \pm0.09 $& $0.01 \pm0.01 $& $0.09 \pm0.04 $ \\
    \midrule
    Random &$0.25 \pm 0.01$&$0.25 \pm 0.01$&$0.24 \pm 0.03$&$0.24 \pm 0.01$&$0.22 \pm 0.01$&$0.24 \pm 0.01$\\
    \bottomrule
  \end{tabular}
  \caption{The valid action rate performance of LLMs on the benchmark levels with standard error.}
\end{table*}

\section{Additional Details on RL Experiments}
\label{appendix:rl_experiment}
Both Ape-X DQN and PPO algorithms use implementations from RLlib \cite{liang2018rllib}, take pixel-based observations as input, and adopt a multiple-layers CNN-based architecture similar to that used in the original DQN paper \cite{mnih2015human}, with approximately 3.1 million parameters, as shown in Figure~\ref{fig:nn}. We also evaluated both algorithms under the structured game-state observation setting, with their model architectures shown in Figure 7. Each algorithm was trained on a personal computer equipped with an Intel(R) Core(TM) i7-14700K CPU (20 cores), an NVIDIA GeForce RTX 3090 GPU with 24 GB of VRAM, and 192 GB of RAM. We applied practicality-oriented environment preprocessing to the TowerMind environment interface for training: (1) we downsampled the original $512 \times 512 \times 3$ pixel-based observations to $128 \times 128 \times 3$; (2) we discretized TowerMind's hybrid action space by uniformly dividing the first two continuous action variables (representing horizontal and vertical coordinates) into 10 discrete intervals ranging from -3.0 to 3.0, while keeping the action type unchanged, resulting in a discrete action space of size $10 \times 10 \times 12$, as shown in Figure~\ref{fig:split}; and (3) we introduced an additional step penalty \cite{sowerby2022designing} of $-5 \times 10^{-4}$ to the original TowerMind reward structure. During RL training, each episode was initialized by randomly selecting one of the nine benchmark levels. See Table~6 for hyperparameters used for training Ape-X DQN and PPO.

\begin{figure}[htbp]
  \centering
  \includegraphics[clip=false, width=\linewidth]{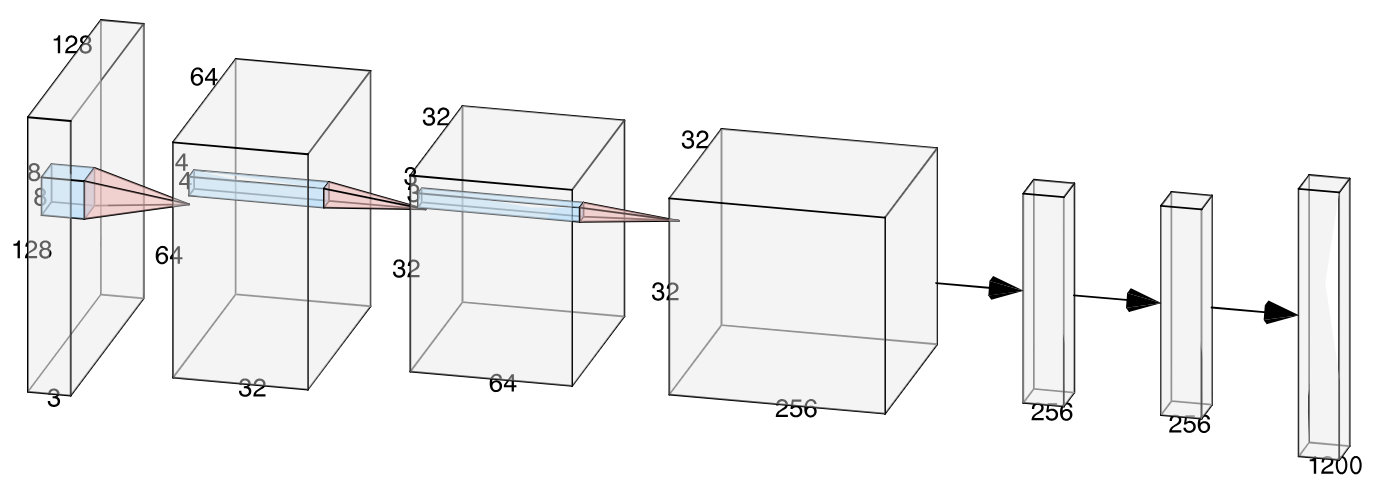}
  \caption{Model architecture of the multiple-layers CNN model, the 128 x 128 × 3 image is processed thorough convolution layers, ultimately producing a 1200 dimensional policy output via an full connection network }
  \label{fig:nn}
\end{figure}

\begin{figure}[htbp]
  \centering
  \includegraphics[clip=false, width=\linewidth]{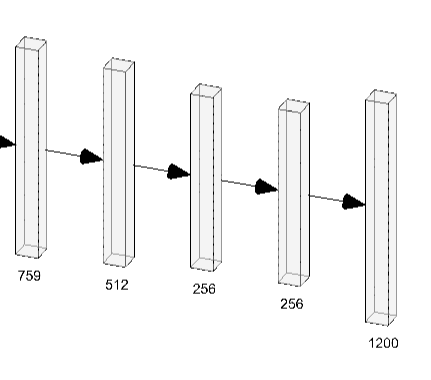}
  \caption{Model architecture of the MLP model, the structured game-state observation (length 759) is fed into fully connected layers, resulting in a 1200-dimensional policy output. }
  \label{fig:nn}
\end{figure}

\begin{figure}[htbp]
  \centering
  \includegraphics[clip=false, width=\linewidth]{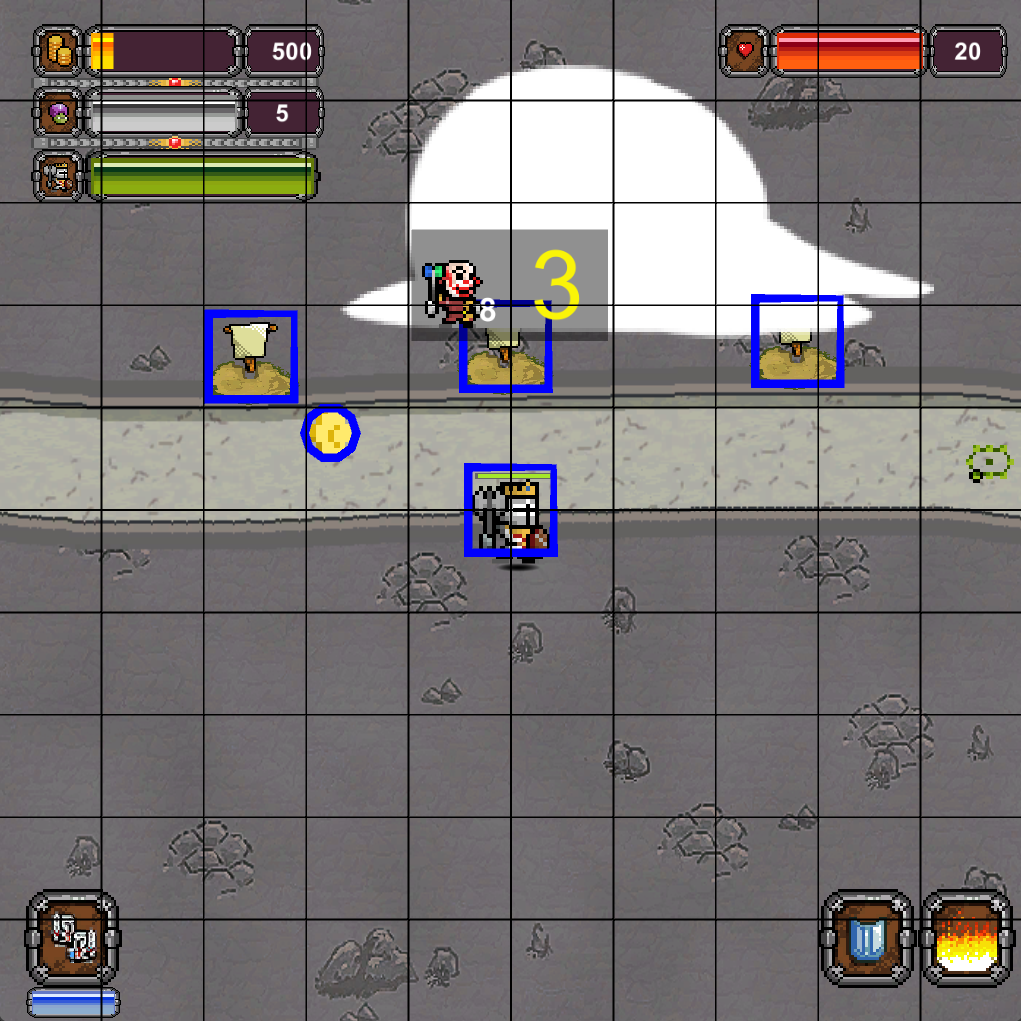}
  \caption{Illustration of action space discretization. The map is divided into a 10 × 10 grid, and actions are performed at the centers of the grid cells instead of using continuous coordinates.}
  \label{fig:split}
\end{figure}

\begin{table}[ht]
  \centering
    \fontsize{9pt}{9pt}\selectfont
  \label{tab:ppo_apex_hparams}
  \small
  \begin{tabular}{@{}lcc@{}}
    \toprule
    \textbf{Parameter}                       & \textbf{Ape-X DQN} & \textbf{PPO}  \\
    \midrule
    Optimizer                                 & Adam                   & Adam \\
    Learning rate                             & $0.0003$ & $0.0003$   \\
    Other optimizer params                    & $\varepsilon = 0.00015$& ---                     \\
    Discount factor ($\gamma$)                & --- & 0.99                   \\
    GAE $\lambda$                            & ---  & 0.95                   \\
    Policy clip ratio                         & ---& 0.20                    \\
    $n$-step returns                          & 3& ---                     \\
    Train batch size                          & 64& 512                     \\
    SGD minibatch size                        & ---& 128                     \\
    Number of SGD iterations                  & ---& 30                      \\
    Training intensity                        & 8 & ---                    \\
    Double-Q                                  & true& ---                     \\
    Target network update freq               & 20\,000 & ---                     \\
    Replay capacity                            & 1500000& ---                    \\
    Prioritized replay $\alpha$               & 0.5& ---                     \\
    Prioritized replay $\beta$                & 1.0 & ---                    \\
    Initial / final $\epsilon$                & 1.0 / 0.01& ---                     \\
    \bottomrule
  \end{tabular}
  \caption{Key hyperparameters for Ape-X DQN and PPO in our experiments.  
           A dash (---) indicates the parameter is not specified or not applicable for the corresponding algorithm.}
\end{table}

\end{document}